\documentclass[journal,onecolumn]{IEEEtran}

\usepackage{epsfig}
\usepackage{graphicx}
\usepackage{amsmath}
\usepackage{amssymb}
\usepackage{commath}
\usepackage{bbding}
\usepackage{cite}
\usepackage{cleveref}
\usepackage{booktabs,multirow}
\usepackage{times}

\usepackage{mathtools}

\begin{document}

\title{Non-Linear Phase-Shifting of Haar Wavelets for Run-Time All-Frequency Lighting}
%
%

\author{Mais~Alnasser\thanks{Mais Alnasser was with the Department of Computer Science, University of Central Florida, Orlando, FL, 32816 USA at the time this project was conducted. (e-mail: nasserm@cs.ucf.edu).} and Hassan~Foroosh\thanks{Hassan Foroosh is with the Department of Computer Science, University of Central Florida, Orlando, FL, 32816 USA (e-mail: foroosh@cs.ucf.edu).}
}

\maketitle

\begin{abstract}
This paper focuses on real-time all-frequency image-based rendering using an innovative solution for run-time computation of light transport. The approach is based on new results derived for non-linear phase shifting in the Haar wavelet domain. Although image-based methods for real-time rendering of dynamic glossy objects have been proposed, they do not truly scale to all possible frequencies and high sampling rates without trading storage, glossiness, or computational time, while varying both lighting and viewpoint. This is due to the fact that current approaches are limited to precomputed radiance transfer (PRT), which is prohibitively expensive in terms of memory requirements and real-time rendering when both varying light and viewpoint changes are required together with high sampling rates for high frequency lighting of glossy material. On the other hand, current methods cannot handle object rotation, which is one of the paramount issues for all PRT methods using wavelets. This latter problem arises because the precomputed data are defined in a global coordinate system and encoded in the wavelet domain, while the object is rotated in a local coordinate system. At the root of all the above problems is the lack of efficient run-time solution to the nontrivial problem of rotating wavelets (a non-linear phase-shift), which we solve in this paper.
\end{abstract}


\section{Introduction}

Image-based rendering (IBR) has been an active area of research in computational imaging and computational photography in the past two decades. It has led to many interesting non-traditional problems in image processing and computer vision, which in turn have benefited from traditional methods such as shape and scene description \cite{Cakmakci_etal_2008,Cakmakci_etal_2008_2,Zhang_etal_2015,Lotfian_Foroosh_2017,Morley_Foroosh2017,Ali-Foroosh2016,Ali-Foroosh2015,Einsele_Foroosh_2015,ali2016character,Cakmakci_etal2008,damkjer2014mesh,Junejo_etal_2013,bhutta2011selective,junejo1dynamic,ashraf2007near,Junejo_etal_2007,Junejo_Foroosh_2008,Sun_etal_2012,junejo2007trajectory,sun2011motion,Ashraf_etal2012,sun2014feature,Junejo_Foroosh2007-1,Junejo_Foroosh2007-2,Junejo_Foroosh2007-3,Junejo_Foroosh2006-1,Junejo_Foroosh2006-2,ashraf2012motion,ashraf2015motion,sun2014should},  scene content modeling \cite{Junejo_etal_2010,Junejo_Foroosh_2010,Junejo_Foroosh_solar2008,Junejo_Foroosh_GPS2008,junejo2006calibrating,junejo2008gps,Tariq_etal_2017,Tariq_etal_2017_2,tariq2013exploiting,tariq2015feature,tariq2014scene}, super-resolution (in particular in 3D) \cite{Hu_etal_IBR2012,Foroosh_2000,Foroosh_Chellappa_1999,Foroosh_etal_1996,Cao_etal_2015,berthod1994reconstruction,shekarforoush19953d,lorette1997super,shekarforoush1998multi,shekarforoush1996super,shekarforoush1995sub,shekarforoush1999conditioning,shekarforoush1998adaptive,berthod1994refining,shekarforoush1998denoising,bhutta2006blind,jain2008super,shekarforoush2000noise,shekarforoush1999super,shekarforoush1998blind}, video content modeling \cite{Shen_Foroosh_2009,Ashraf_etal_2014,Ashraf_etal_2013,Sun_etal_2015,shen2008view,sun2011action,ashraf2014view,shen2008action,shen2008view-2,ashraf2013view,ashraf2010view,boyraz122014action,Shen_Foroosh_FR2008,Shen_Foroosh_pose2008,ashraf2012human},  image alignment \cite{Foroosh_etal_2002,Foroosh_2005,Balci_Foroosh_2006,Balci_Foroosh_2006_2,Alnasser_Foroosh_2008,Atalay_Foroosh_2017,Atalay_Foroosh_2017-2,shekarforoush1996subpixel,foroosh2004sub,shekarforoush1995subpixel,balci2005inferring,balci2005estimating,foroosh2003motion,Balci_Foroosh_phase2005,Foroosh_Balci_2004,foroosh2001closed,shekarforoush2000multifractal,balci2006subpixel,balci2006alignment,foroosh2004adaptive,foroosh2003adaptive}, tracking and object pose estimation \cite{Shu_etal_2016,Milikan_etal_2017,Millikan_etal2015,shekarforoush2000multi,millikan2015initialized}, and camera motion quantification and calibration \cite{Cao_Foroosh_2007,Cao_Foroosh_2006,Cao_etal_2006,Junejo_etal_2011,cao2004camera,cao2004simple,caometrology,junejo2006dissecting,junejo2007robust,cao2006self,foroosh2005self,junejo2006robust,Junejo_Foroosh_calib2008,Junejo_Foroosh_PTZ2008,Junejo_Foroosh_SolCalib2008,Ashraf_Foroosh_2008,Junejo_Foroosh_Givens2008,Lu_Foroosh2006,Balci_Foroosh_metro2005,Cao_Foroosh_calib2004,Cao_Foroosh_calib2004,cao2006camera}, to name a few.

Using images to estimate or model environment light for relighting objects introduced or rendered in a scene is a central problem in this area \cite{Cao_etal_2005,Cao_etal_2009,shen2006video,balci2006real,xiao20063d,moore2008learning,alnasser2006image,Alnasser_Foroosh_rend2006,fu2004expression,balci2006image,xiao2006new,cao2006synthesizing}.
This requires solving the light integral equation (also known as the rendering equation), which plays a crucial role in IBR. IBR and new view synthesis has been the subject of intensive research for the past two decades, motivated by various application areas, such as interactive entertainment (e.g. video games), 3D TV, and augmented reality, to name a few. Image-based view synthesis techniques can be broadly classified into two categories. Those that do not require 3D information, such as plenoptic methods and light-field rendering techniques \cite{Do12,Kubota07,McMillan95}, and those that use some 3D information either implicitly or explicitly, such as the depth image-based rendering (DIBR) methods \cite{Koppel10,Koppel10-2,Lee09,Maugey12,Min10,Nguyen06,Pearson11}. A specific problem of interest in this area is also when it is desired to render a 3D object of interest within an environment captured by 2D image(s) - either a single image or several images like in plenoptic data. The latter methods may be dealing with only the relighting problem (image-based relighting) when the viewpoint is assumed stationary but the light changing dynamically, or as a more complex rendering and relighting problem, when the viewpoint is also changing due to translations and rotations \cite{Agarwal03Structured,Kollig03,Ng03AllFreqShadow,Ng04Triple,Ostromoukhov04,Pharr04,RammamoorthiEfficientRep}. This latter category of problems may be referred to as Physics Image Based Rendering techniques (PIBR), since they start by modeling the problem as that of solving the light transport equation. A major issue in many of these IBR problems is the amount of storage and bandwidth needed to handle the problem of view synthesis. Most methods in all the categories defined above are based on some level of precomputed information, which often requires also dealing with various issues related to the coding problem \cite{Cheung11,Do12,Maitre08} to tackle storage and bandwidth.

The demand for photorealism in image-based rendering and relighting has been increasing in all the various categories of methods highlighted above. Parallel to photorealism, coding and compression have been extensively studied and tackled primarily as off-line issues. In this paper, we are proposing a run-time solution to PIBR that tackles the bandwidth and storage issues by solving a non-linear phase-shifting problem in the Haar domain. As such, since memory and bandwidth requirements are addressed at the run time, our solution provides a means of reducing these requirements by trading them against computation during the rendering process. Furthermore, our solution provides this advantage by directly working in the compressed domain (Haar domain), thereby making it possible to achieve the above goal with little computational overhead. In order to better understand the challenges we first review the methods in the area of PIBR.

Early attempts to PIBR have been mainly concerned with realistic rendering. Phong \cite{Phong75}, Cook and Torrance \cite{Cook81RefModel}, and  Blinn \cite{Blinn98} generated glossy highlights from point sources by using general specular reflection functions that are concentrated near the mirror direction. Cook, Carpenter and Porter \cite{Cook88DrayTracing} used distributed ray tracing to
model glossy reflections for a specular reflection function and an
arbitrary light distribution. Blinn and Newell were the first to
include the light coming from the whole environment using
environment maps to create mirror reflections
\cite{Blinn76Reflection}. Kajiya \cite{Kajiya86RenderingEq}
introduced the \emph{Rendering Equation}, which provided a unified
framework for rendering different types of materials and became thereafter the focus of research for the rendering community.

There has been a remarkable progress in the last few years in finding efficient ways of solving the PIBR problem. The most recent approaches have been focusing on projecting the problem into the frequency domain such as the Spherical Harmonics or the Haar transform domain. PIBR techniques that use Haar wavelets for representation rely highly on preprocessing of data. Ng et. al. \cite{Ng04Triple} discretize BRDFs along the possible orientations of the normal direction. This amounts to 6-dimensional BRDFs and gigabytes of storage. Wang et. al. \cite{Wang06WaveRot} provide an alternative solution to rotating Haar-transformed data using geometry maps. They preprocess and store the rotation matrices instead of the BRDFs. However, their method relies on discretizing the space of possible normal orientations and interpolating between the rotation matrices at those sample directions. We provide an analytic solution that offers the option of computing rotated Haar-transformed data directly, using only the high frequency coefficients of the original data without having to resort to discretization, storage, interpolation of data, or trading off the bandwidth.

Computer systems and Graphic cards are becoming faster and more powerful. They are also becoming increasingly more capable in terms of memory. In rendering applications that favor real or interactive time, memory is often traded for speed. This essentially relies on preprocessing, which in turn relies on discretizing the data as finely as the memory would allow. However, no matter how large the size of the memory, there is always a limit to the resolution of discretization, beyond which one has to approximate the values between samples using interpolation. This is the natural effect of discretization. At the rate the speed of systems is increasing, applications have started favoring adding more computation during render time to achieve more precision - a trend in the graphics hardware community that also favors our solution for the PIBR problem.

Our run-time rendering and relighting solution requires rotating Haar wavelets, which amounts to a non-linear phase-shifting directly in the Haar wavelet domain. In a previous work \cite{Alnasser2D}, we proposed a solution for linear phase-shifting of multi-dimensional non-separable Haar wavelets. In this paper, we present an innovative method for non-linear phase-shifting that takes advantage of our earlier results in \cite{Alnasser2D}. This is possible since the non-linear phase-shifting of the two-dimensional signals in our case are induced by linear phase shifts of three-dimensional functions defined on a unit sphere. In other words, the non-linear phase-shifting in the two-dimensional signal is a rotation of the three-dimensional data it represents.

In the remaining of this paper, we first provide a brief introduction to PIBR followed by a review of some of the most recent research in this area that are related to our work. We then present a summary of our contributions, followed by a detailed description of non-linear phase-shifting of the 2D non-separable Haar wavelets and its applications to the problem at hand. We finally provide experimental results and conclude with some remarks and discussion.

\section{PHYSICS IMAGE-BASED RENDERING}
As pointed out earlier, PIBR methods are based on modeling the physics of the rendering problem using the light transport integral, also known as the rendering equation. The rendering equation describes the interaction between incoming
light and a surface material. The following is the most general form
of the equation:
\begin{eqnarray}
L_{o}(p,\omega_{o}) \!\!\!\!\!\!\!\!&=&\!\!\!\!\!\!\!\! L_e(p,\omega_o) \nonumber \\
\!\!\!\!\!\!\!\!&+&\!\!\!\!\!\!\!\!\int_{H(\vec{N})}f_r(p,\omega_o,\omega_{in}) L_{in}(p,\omega_{in})
V(p,\omega_{in})\cos\theta_{in}{\tt d}\omega_{in}\nonumber \\
\end{eqnarray}
where, \begin{itemize}
\item \textbf{$\omega_o$} is the outgoing
direction.

\item \textbf{$L_e(p,\omega_o)$} is the emitted light at point $p$ in the direction $\omega_o$.

\item \textbf{$L_o(p,\omega_o)$} is the radiance leaving the surface at a point $p$ in the direction $\omega_o$.

\item \textbf{$f(p,\omega_o,\omega_{in})$} is the Bidirectional
Reflectance Distribution Function (BRDF).

\item \textbf{$L_{in}(p,\omega_{in})$} is the incident radiance.

\item \textbf{$\theta_{in}$} is the angle between the unit normal
$\vec{N}$ at the point $p$ and $\omega_{in}$.

\item \textbf{$V(p, \omega_{in})$} is the function that describes
the visibility at point $p$ along direction $\omega_{in}$.

\item \textbf{$H(\vec{N})$} is the hemisphere of directions around the normal
$\vec{N}$.
\end{itemize}

\begin{figure}[h]
\centering
\includegraphics[width=3in]{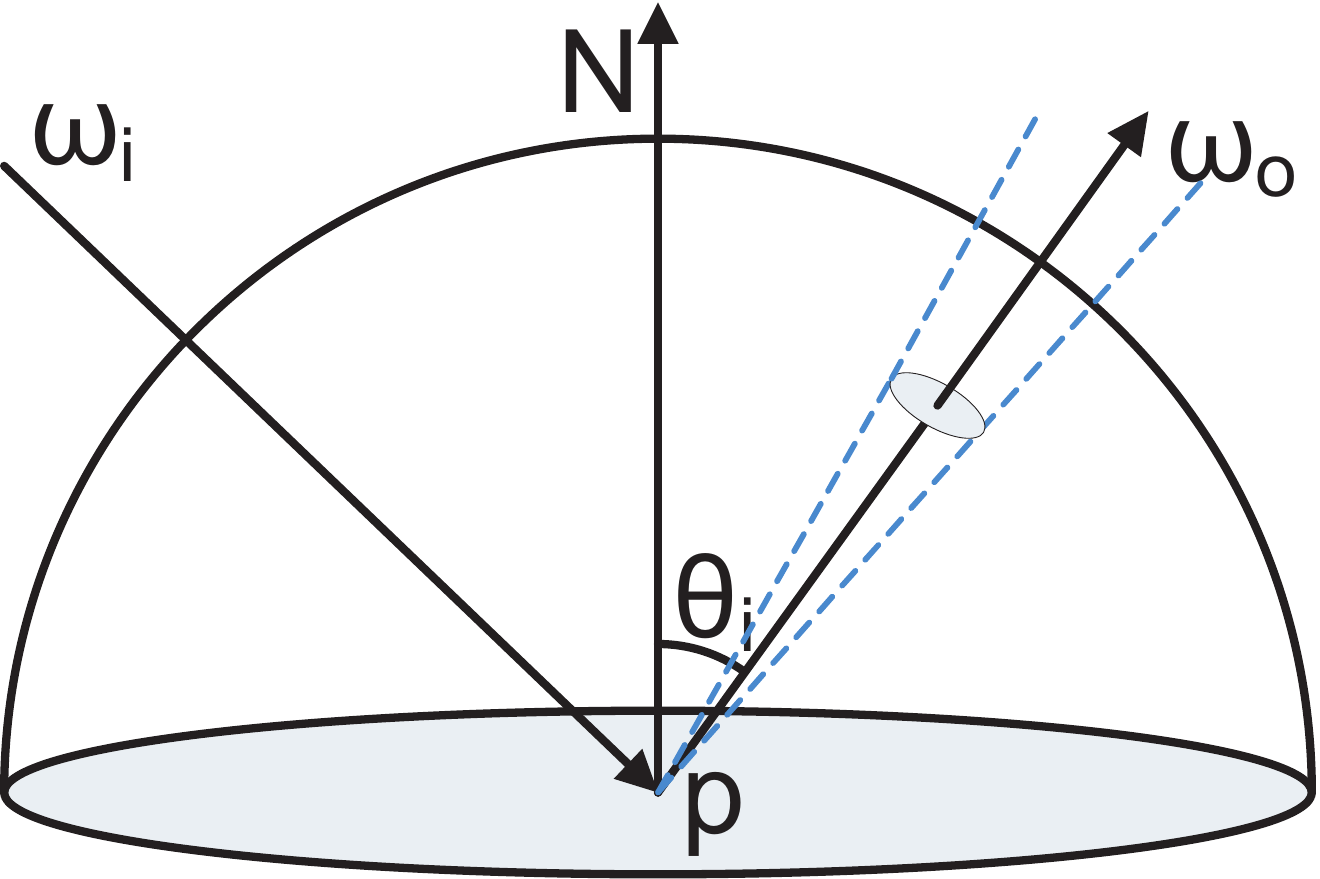}
\caption{Relationship between the incident light and the light
leaving a point on an object surface.} \label{fig:light}
\end{figure}

Put in words, the rendering equation integrates the product of the
three contributing factors of incoming light, BRDF, and the
visibility function along all incident light directions. Figure
\ref{fig:light} shows the relationship between the incident light
and the reflected light off the surface of an object at a given
point.

\section{CLOSELY RELATED WORKS}
We describe below three categories of work that are closely related to our paper, most notably the methods working in the transform domain.
\subsection{MONTE CARLO}
One of the oldest and most straightforward approaches for solving
the light integral equation is to approximate the solution using the
Monte Carlo method ~\cite{Kalos86MC,Kajiya86RenderingEq,Dutre02}.
Global illumination algorithms based on Monte Carlo are general
enough to allow estimation of the integral for any material type and
light distribution.

Monte Carlo is a numerical integration method that uses sampling
to estimate an average solution for integration of any dimension.
This is applicable to the lighting integral because the product of
the light, reflection and visibility functions is too complex to
evaluate using a closed-form approach.

Monte Carlo based algorithms are, however, very slow. The
convergence rate for these algorithms is
\emph{O}($\frac{1}{\sqrt{n}}$), where $n$ is the number of samples
taken to estimate the integral. This means that to cut the error
in half, four times the number of samples must be taken. On the
other hand, unless sufficient light samples are taken, Monte Carlo
produces noisy results that manifest as pixels that are too bright
or too dark in the rendered image. Therefore, a substantial number of samples and
accordingly more time is typically required in order to render a
realistic low-noise image.

Much research has gone into improving Monte Carlo's performance
without increasing the number of samples. One of the
techniques that has been most effective is importance sampling
\cite{Agarwal03Structured,Kollig03,Ostromoukhov04}. Importance
sampling relies on sampling mostly in the ``important''
directions, which is governed by the choice of a sampling
distribution function that is similar in shape to the integrand of
the function that is being estimated \cite{Pharr04}.

Many attempts have been also made to speed the rendering time by
splitting the scene synthesis into an offline prefiltering
preprocess and a rendering process. Prefiltering stores the result
of integrating the product of the BRDF and lighting over the
visible upper hemisphere per normal direction. Cabral et al.
\cite{Cabral99ReflSpace} used prefiltering to obtain a sparse 2D
set of prerendered images that were used during the rendering
process to generate rendered images at interactive rates.
\cite{kautz00approximation} and \cite{Kautz00unified} subsequently
proposed alternative methods for improving prefiltering methods.

\subsection{FREQUENCY DOMAIN}
\subsection{SPHERICAL HARMONICS}
The first attempt to solve the integral in the frequency domain was by
Cabral et al. \cite{Cabral87BRDFbumpMaps} using \emph{Spherical
Harmonics} as basis. Spherical harmonics are the analog of the
Fourier transform for representing functions on the unit sphere
\cite{MacRobert48SH}. They are the products of \emph{Associated
Legendre Functions} with functions that are periodic with respect
to the azimuth angle $\phi$. 

Cabral et al. \cite{Cabral87BRDFbumpMaps} simplified the integral
by removing the emittance and the visibility functions. They also
made the assumptions that the viewing direction is fixed and that
the BRDF is isotropic. They then used spherical harmonics to
expand the lighting function and the product of the BRDF and the
cosine function, with the viewing direction as the north pole.
Projecting the terms of the integral into the spherical harmonics
space reduces the integration into an inner product because of the
orthonormality of spherical harmonics. 

The next use of spherical harmonics was by Sillion et al.
\cite{Sillion91GIrefDist} to represent exit radiance at many
points in a progressive radiosity simulation. They provide a
general off-line treatment of transfer functions ranging from
ideal specular to ideal diffuse reflection.

\begin{figure}[h!]
    \begin{center}
    \includegraphics[width=3in,height=2.7in]{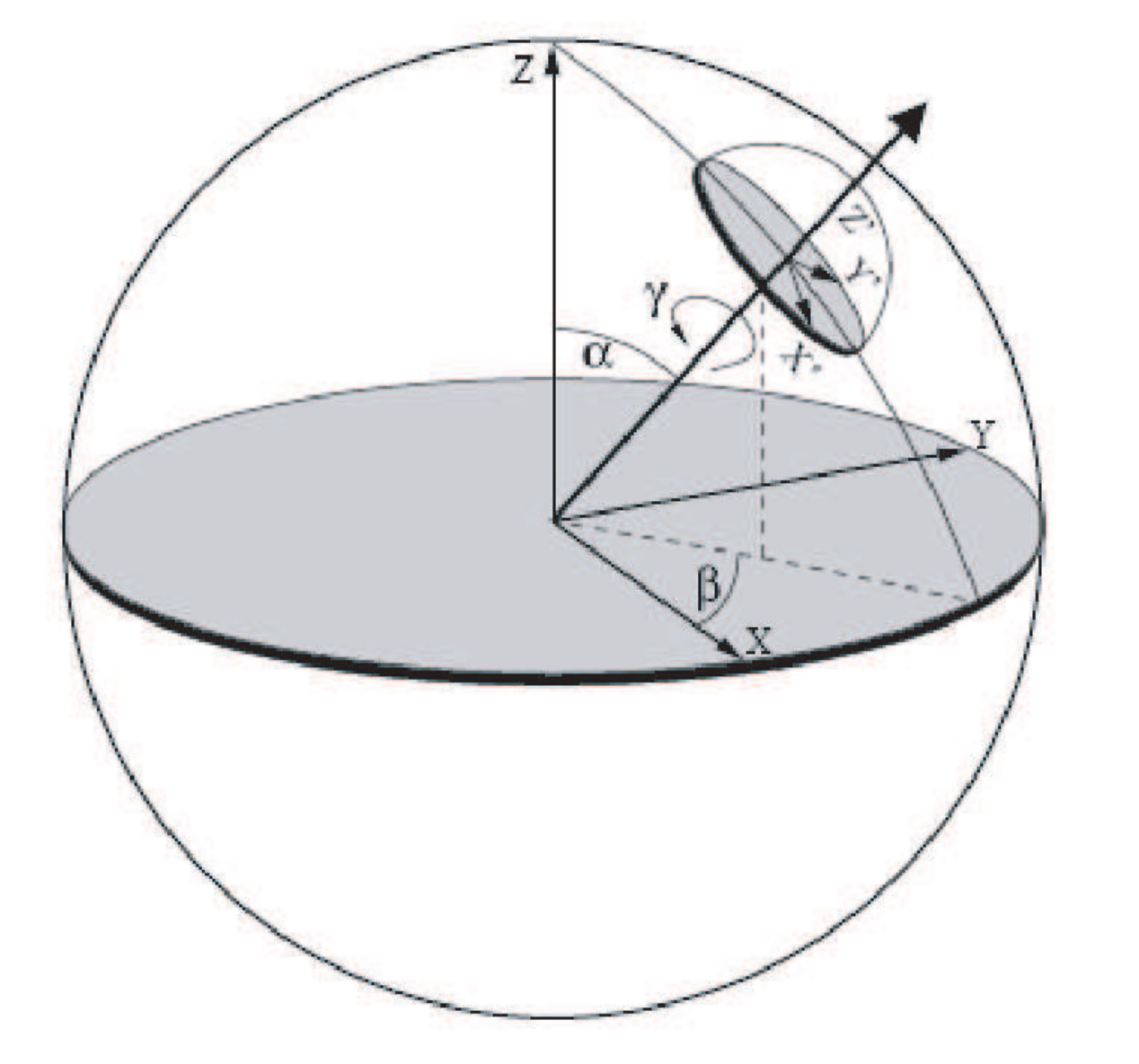}
    \end{center}
    \caption{The local and global coordinate systems.} \label{fig:locl}
\end{figure}

Ramamoorthi and Hanrahan \cite{RammamoorthiEfficientRep} gave
explicit formulae in terms of the cartesian coordinates of the
surface normal and approximated the solution of the integral for
diffuse materials using 9 coefficients only. Their solution was
the first to use spherical harmonics for realistic real-time
rendering in the area of PIBR. Their formulae implicitly handle the required rotation
of light coefficients into the local space defined by the normal
as its Y-direction. This rotation is required because of the
different coordinate spaces of the BRDF and the light function.
The BRDF is sampled and projected into spherical harmonics'
frequency domain in local space, that is, the surface normal is
the Y-axis of the space. On the other hand, the light is sampled
and projected in the global space, that is, the up-direction is
the Y-axis of the space (see Fig. \ref{fig:locl}). Kautz et al. \cite{Kautz02SH} extend the
use of spherical harmonics to arbitrary BRDF's under low-frequency
lighting. They represented the lighting environment using 25
coefficients and rotated them for each vertex during
rendering at interactive rates. They also combined their method
with Precomputed Radiance Transfer (PRT) \cite{Sloan02PRT} to handle
interreflections and shadows. Functions represented by spherical
harmonics can be rotated by a linear transformation of the
representation coefficients \cite{Green03Gritty}. The existing
procedures
\cite{Ivanic96Rotation,Ivanic98Rotation,Choi99Rotation,Kautz02SH},
however, are slow and cause a bottleneck in the rendering process.
K\v{r}iv\'{a}nek et al. \cite{Krivanek06} propose an efficient
approximation of the spherical harmonic rotation based on replacing
the general spherical harmonic rotation matrix with its truncated
Taylor expansion. Their proposed rotation approximation is faster
and has a lower computational complexity. The approximation,
however, is accurate only for small rotation angles and, therefore,
applicable only to certain applications that require small
successive rotation angles.

Spherical harmonics are shift-invariant, which makes them suitable
for representing functions on the sphere. However, They are
globally supported and, therefore, suffer from some of the same
difficulties as the Fourier transform on the line such as the Gibbs phenomenon.
Furthermore, spherical harmonics don't have good localization,
which means a large number of coefficients is required for
representing high frequency functions. Also, to handle the
visibility function, it has to be combined with the BRDF and
projected as one function, referred to as the \emph{Transfer
Function}, in order to be able to convert the lighting integral
into an inner product. This restricts the BRDF representation in
the sense that rather than having one representation for a certain
material, there would be several representations of the same
material, each dependant on the object occlusion properties.
Furthermore, the redundancy of the BRDF information occurs per
sample point per object due to the dependence of occlusion on the
position of the sample point.

Imposing constraints to simplify the integral to achieve real-time
or interactive rates is a necessity with spherical harmonics, when
glossy or general BRDF's are used. One method is to use low sampling
rates, which is equivalent to band-limiting the illumination. This
approach is used by Sloan et al. in \cite{Sloan02PRT},
\cite{Sloan03BiScaleRT} and \cite{Sloan03PCA}. However,
band-limiting removes high frequency components, which blurs
lighting detail. For diffuse materials, the error can be very low
\cite{RammamoorthiEfficientRep}. However, this approach is not
suitable for glossy materials, which need the high frequency for an
efficient representation. Another approach is to reduce the
dimensionality of the problem by fixing the light or the viewing
direction as in Kautz et al. \cite{Kautz02SH}. This approach,
however, restricts the dynamics of the scene.

\subsection{HAAR-DOMAIN METHODS}
Ng et al. \cite{Ng03AllFreqShadow} used non-linear wavelet
approximation to achieve better localization than spherical
harmonics and were successful at representing different degrees of
shadowing due to wavelets' excellent capability in handling
information at different scales. They, however, reduced the
dimensionality of the integral to simplify it by fixing the
viewing direction. Ng et al. \cite{Ng04Triple} developed a method
for solving the triple-product integral using two-dimensional
non-separable Haar wavelets to avoid reducing the dimensionality
of the problem and the low sampling rate inherent in spherical
harmonics methods. 
to a triple product:

\begin{eqnarray}
L_{o}(p,\omega_{o}) & = &
\int_{S}L_{in}(p,\omega_{in})f_r(p,\omega_o,\omega_{in})
\cos\hspace*{-0.5mm}\theta_{in}V(p,\omega_{in})\hspace*{0.5mm}d\hspace*{-0.5mm}\omega_{in}\\
& = &
\int_{S}L_{in}(p,\omega_{in})\tilde{f}_r(p,\omega_o,\omega_{in})V(p,\omega_{in})
\hspace*{0.5mm}d\hspace*{-0.5mm}\omega_{in}\\
& \approx &
\int_{S}(\sum_ia_i(p)\Psi_i(\omega_{in}))(\sum_jb_j(p,\omega_o)\Psi_j(\omega_{in}))
(\sum_kc_k(p)\Psi_k(\omega_{in}))d\hspace*{-0.5mm}\omega_{in}\\
& = & \sum_i\sum_j\sum_k a_i(p) b_j(p,\omega_o) c_k(p)
\int_{S}\Psi_i(\omega_{in})\Psi_j(\omega_{in})\Psi_k(\omega_{in})d\hspace*{-0.5mm}\omega_{in}\\
& = & \sum_i\sum_j\sum_k C_{ijk} a_i(p) b_j(p,\omega_o)
c_k(p)\label{eq:tripleSum}
\end{eqnarray}
where,
\begin{equation*}
C_{ijk} =
\int_{S}\Psi_i(\omega_{in})\Psi_j(\omega_{in})\Psi_k(\omega_{in})d\hspace*{-0.5mm}\omega_{in}
\end{equation*}

Rhis formulation was general and worked for any basis. However,
solving for $C_{ijk}$ and efficiently computing the triple sum in
(\ref{eq:tripleSum}) was not a trivial matter. One can refer to
\cite{Ng04Triple} for an in-depth analysis of the computational
complexity for the different methods to solve the triple sum.

Haar wavelets provide an efficient solution for solving the triple
sum due to the following simple theorem \cite{Ng04Triple}:
\begin{enumerate}
\item {\em The Tripling Coefficient Theorem} The integral of three 2D
Haar basis functions is non-zero if and only if one of the following
three cases hold:
\begin{enumerate}
\item All three are scaling functions, that is, $C_{ijk}=1$.
\item All three occupy the same wavelet square and all are different
wavelet types, that is $C_{ijk}=2^l$, where wavelets are at level
$l$.
\item Two are identical wavelets, and the third is either the
scaling function or a wavelet that overlaps at a strictly coarser
level, that is, $C_{ijk} = \pm2^l$, where the third function exists
at level $l$.
\end{enumerate}
\end{enumerate}

The above theorem implies that most of the tripling coefficients
are equal to zero because most pairs of basis functions do not
overlap, which is the reason for the efficiency of using 2D Haar wavelets
to solve the triple product integral.

However, Haar wavelets are not rotation-invariant, therefore, the
BRDF has to be sampled per normal direction, which is storage and bandwidth
intensive and has a limit on how much resolution it can afford.
Furthermore, Haar wavelets are parameterized in cube domains to
represent spherical functions. This makes it difficult or impossible
to obtain an analytic rotation formula in the frequency domain. This
is due to the representation mechanism, which compresses each face
of the cube as a separate entity. To be able to rotate in the
frequency domain data has to move from one face to another, which is
not feasible with the cube representation.

Schr\"{o}der and Sweldens \cite{Schroder95sphWavelets} constructed
biorthogonal wavelets on the sphere using the lifting scheme. Their
method achieved better localization than spherical harmonics for
high-frequency glossy materials and was defined directly on the sphere.
However, it was also not rotation-invariant and did not lend itself
to solving the rendering triple integral equation in an efficient manner.

Wang et al. \cite{Wang06WaveRot} parameterized the spherical functions
using geometry maps \cite{Praun03Paramterization} and provided a
solution for wavelet rotation using precomputed rotation matrices.
They precomputed and stored the rotation matrices and used them to
rotate the light coefficients into the local space of the BRDF.
However, their solution was brute force rather than analytic, since it
would be impossible to derive rotation formulae for the frequency
domain with the geometry map representation.

\section{OUR CONTRIBUTIONS}
 \textbf{Rotation directly in the Haar wavelet domain:} We derive for the first time an explicit solution for rotating functions directly in the Haar wavelet domain. The key idea that allows us to achieve this result is the fact that for a standard non-separable Haar transform the horizontal, vertical, and diagonal coefficients are simply first order finite difference approximations of horizontal, vertical, and diagonal derivatives of the function at different scales. Therefore, we first derive the explicit expressions in the spatial domain that describe rotations of a function defined over the unit sphere. Using the chain rule, and the fact that differentiation is a linear operator, we then show how the order of rotation and differentiation can be interchanged. As a result, we derive an explicit method for rotating Haar wavelets, which is essentially a non-linear phase shift in the Haar domain.

\textbf{Scalable solution for light transport:} Direct Haar domain rotation completely removes the precomputation burden and the overly expensive storage and bandwidth requirements, since the rotations between local and global coordinates for the BRDF, light, or visibility can be performed directly on the wavelet coefficients during run-time. Throughout this paper the term ``run-time'' refers to computation of the radiance transfer without performing any precomputation. We thus call our approach a run-time radiance transfer (RRT) method as opposed to PRT described above. Although, our approach provides a run-time solution, as seen below, the first algorithm derived tends to be expensive. Fortunately, however, we demonstrate that a simple reformulation of the algorithm reduces the time complexity drastically. As a result, we obtain a solution that provides run-time rotation of the Haar coefficients over the entire frequency range, without requiring any data-loss, precomputation, or extensive storage or bandwidth requirements: the rotated coefficients for the entire frequency range are computed recursively from only one level of coefficients of the Haar transform of the original map prior to rotation. Therefore, our solution does not have to sacrifice glossiness, against storage and bandwidth. We also don't have to interpolate between preprocessed data, whether it is the data itself or the rotation matrices, which means our method is more accurate at any given rotation angle. This is important especially with high frequency information because interpolation acts as a low-pass filter.

\section{OUR METHOD}

As mentioned earlier, the BRDF and the light function are
represented under different coordinate spaces. The BRDF is sampled
and projected into frequency domain in local space, that is, the
surface normal is the Y-axis of the space. On the other hand, the
light is sampled and projected in global space, that is, the
up-direction is the Y-axis of the space. This requires a rotation
of one of the coordinate spaces into the other before the triple
product can be performed. Currently, the most efficient
computation of the triple product is performed in the Haar domain
\cite{Ng04Triple}. Thus the key to solve the problem is to devise a
method for rotating directly in the Haar domain, which we describe
in the next section.

\subsection{ROTATING HAAR} \label{sec:rot}
We describe our solution for the light. But, it equally applies to any function defined over the unit sphere such as the visibility map or the BRDF. Our implementation, however, rotates the BRDF data from local space into global space due to its higher degree of smoothness when mapped to a two-dimensional square, which reduces errors incurred by rotation.



\begin{figure*}[t!]
    \begin{center}
    \begin{tabular}{ccc}
    \includegraphics[width=2in,height=2in]{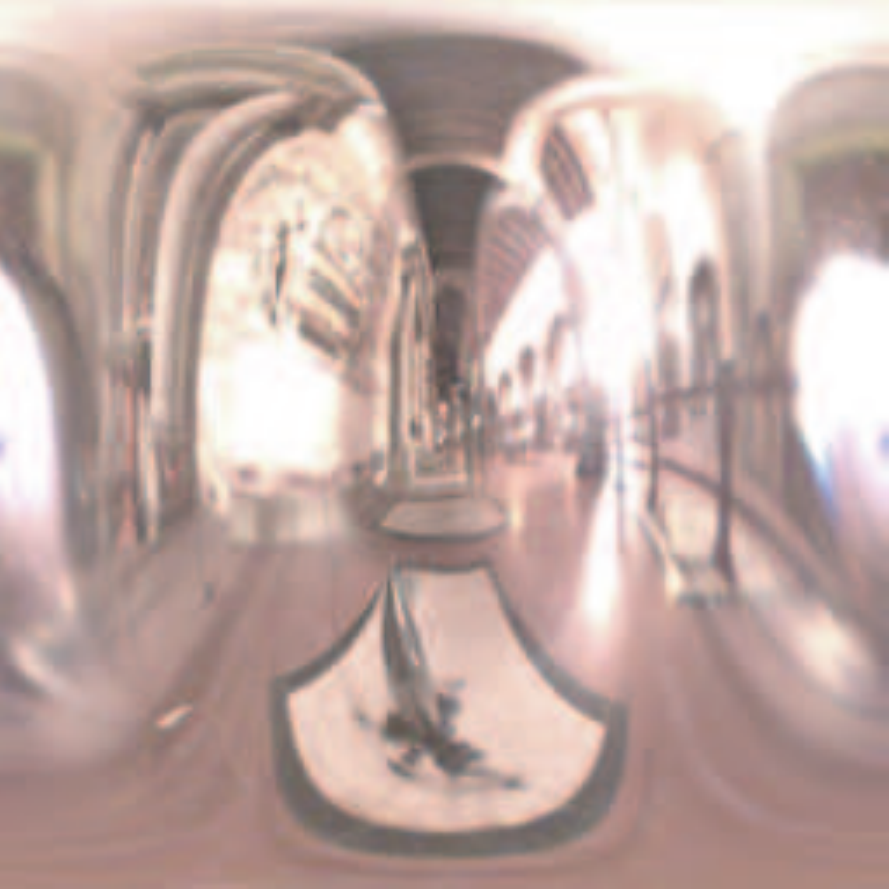}&&
    \includegraphics[width=2in,height=2in]{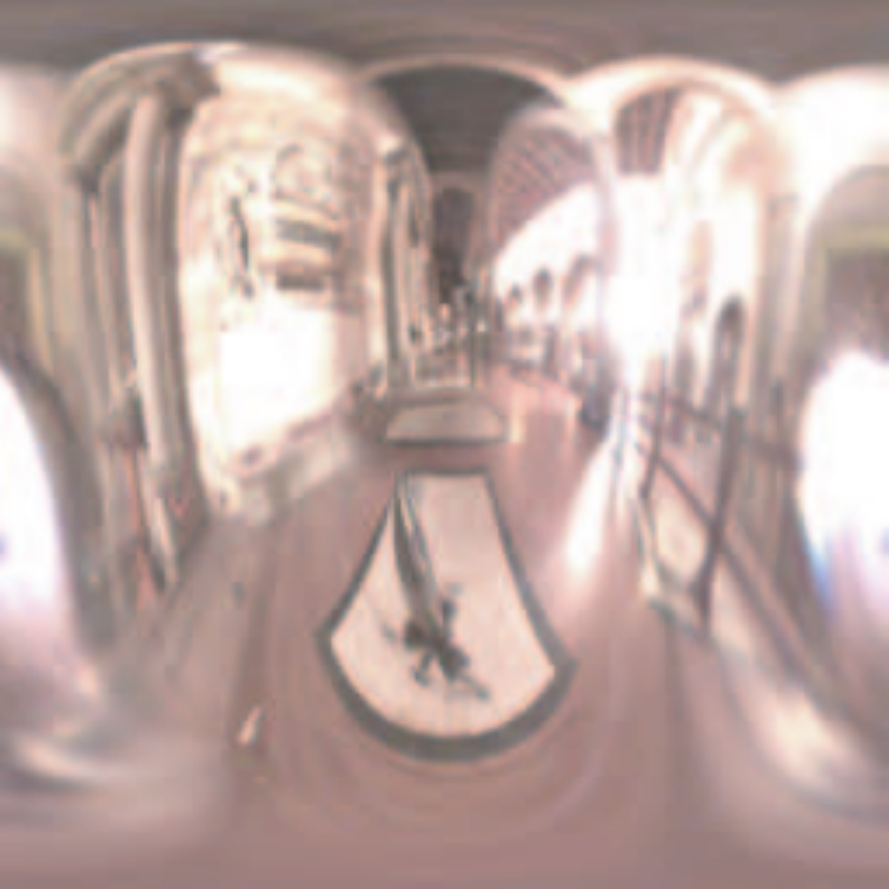}\\
    Original image $f(\theta,\phi)$ && Image after rotation $g(\theta,\phi)$
    \end{tabular}
    \end{center}
        \caption{An example demonstrating a non-linear phase shift that corresponds to a rotation on the unit sphere.} \label{fig:LightMaps}
\end{figure*}

Let $f(\theta,\phi)$ be a function that describes the light map in a spherical coordinate system at some initial orientation\footnote{We do not explicitly include the radius in spherical coordinates to imply that we are dealing with mappings over a unit sphere.}. Our first goal here is to derive the relations that describe rotation in the spherical coordinate system (see Fig. \ref{fig:LightMaps} for an example). Let $\alpha_l$, $\beta_l$ and $\gamma_l$ denote the Euler rotation angles from global to local space along the cartesian axes $X$, $Y$, and $Z$, respectively. Then, the light map after rotation, denoted hereafter by $g(\theta,\phi)$, can be derived in terms of $f(\theta,\phi)$ as follows:

Let $\textbf{p}$ denote a point in the cartesian coordinate system after rotation, i.e. $\textbf{p} \in g(\theta,\phi)$. Then the mapping from spherical coordinates to cartesian coordinates is given by:
\begin{equation}
\textbf{p}=\left[\begin{array}{c}
x \\
y\\
z
\end{array}\right]
=\left[\begin{array}{c}
\sin\theta \sin\phi \nonumber\\
\cos\theta \nonumber\\
\sin\theta \cos\phi
\end{array}\right]
\end{equation}
where we assume a right-handed global coordinate system, with the $Y$-axis pointing up and the $Z$-axis representing the depth. The position of the point prior to an arbitrary rotation $\textbf{R} \in f(\theta,\phi)$ is determined by the following matrix equation:
\begin{equation}
\textbf{p}'=\textbf{R}_z \textbf{R}_y \textbf{R}_x \textbf{p}=\textbf{R} \textbf{p}
\end{equation}
where $\textbf{R}_x$, $\textbf{R}_y$ and $\textbf{R}_z$ are the familiar rotation matrices along the corresponding $X$, $Y$ and $Z$ axes.
For brevity, we can write the elements of $\textbf{p}'$ as follows:
\begin{equation}
\textbf{p}'= \left[\begin{array}{c}
\textbf{R}_1\textbf{p} \\ \\\textbf{R}_2\textbf{p} \\ \\\textbf{R}_3 \textbf{p}
\end{array}
\right] \label{eq:tRot}
\end{equation}
where $\textbf{R}_i, i=1,...,3$ are the rows of the rotation matrix $\textbf{R}$.

Assuming that the point prior to rotation has an elevation angle of
$\theta'$ and an azimuth angle of $\phi'$, we can readily verify that
\begin{equation}
g(\theta,\phi) =
f(\theta',\phi')\nonumber
\end{equation}
where
\begin{eqnarray}
\theta' & = & \cos^{-1}\left(\textbf{R}_2\textbf{p}\right) \label{eq:theta}\\
\phi' & = &
\tan^{-1}\left(\frac{\textbf{R}_1\textbf{p}}{\textbf{R}_3 \textbf{p}}\right) \label{eq:phi}
\end{eqnarray}

Unfortunately, one can easily see from the above relations that the rotation is not a linear operation in the spherical coordinate system. In other words, $\theta'$ and $\phi'$ are not linearly related to $\theta$ and $\phi$. The key observation that allows us to solve the problem is that we are specifically dealing with Haar transform coefficients, which correspond to the first order finite difference approximations of the horizontal, vertical and diagonal derivatives of the map $f(\theta,\phi)$ at different resolution levels (scales).

Therefore, let $\theta'=\Theta(\theta,\phi)$ and $\phi'=\Phi(\theta,\phi)$, denote the non-linear mappings that relate the elevation and azimuth angles before and after rotation, as given by (\ref{eq:theta}) and (\ref{eq:phi}). Then
\begin{equation}
g(\theta,\phi) = f(\Theta(\theta,\phi),\Phi(\theta,\phi)) \label{eq:compfunc}
\end{equation}
The horizontal, vertical and diagonal derivatives are essentially
the derivatives with respect to the angles $\phi$, $\theta$ and $\phi\theta$, respectively. The illuminating feature of (\ref{eq:compfunc}) is that it indicates that the solution to our problem lies simply in the chain rule:
\begin{eqnarray}
\frac{\mbox{d}g}{\mbox{d}\theta} & = &
\frac{\partial{f}}{\partial\Theta}
\frac{\partial\Theta}{\partial\theta} +
\frac{\partial{f}}{\partial\Phi} \frac{\partial\Phi}{\partial\theta}
\label{eq:pde1}\\
\frac{\mbox{d}g}{\mbox{d}\phi} & = &
\frac{\partial{f}}{\partial\Theta}
\frac{\partial\Theta}{\partial\phi} +
\frac{\partial{f}}{\partial\Phi} \frac{\partial\Phi}{\partial\phi} \label{eq:pde2}
\\
\frac{\mbox{d}^2g}{\mbox{d}\phi\mbox{d}\theta} & = &
\frac{\partial^2{f}}{\partial\Theta\partial\theta}
\frac{\partial\Theta}{\partial\theta} +
\frac{\partial^2{f}}{\partial\Theta}
\frac{\partial^2\Theta}{\partial\phi\partial\theta} \nonumber \\
&+&
\frac{\partial^2{f}}{\partial\Phi\partial\theta}
\frac{\partial\Phi}{\partial\phi} + \frac{\partial{f}}{\partial\Phi}
\frac{\partial^2\Phi}{\partial\phi\partial\theta}\label{eq:pde3}
\end{eqnarray}

\begin{figure*}[t!]
\includegraphics[width=6.9in,height=1.5in]{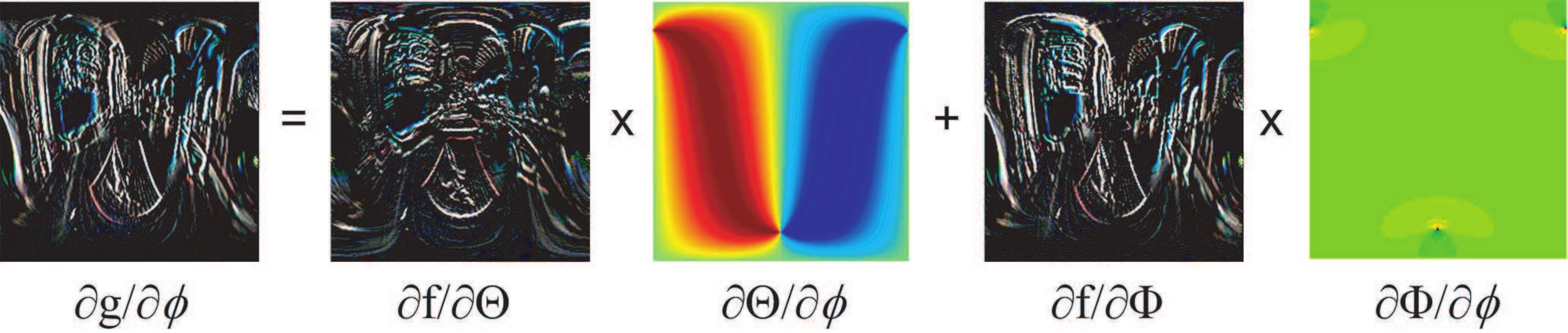}
\caption{This figure shows an example of evaluating
$\frac{\mbox{\small{d\textsl{g}}}}{\mbox{\small{d}}\phi}$ in
equation (\ref{eq:pde2}) after a rotation along the elevation angle by
$20^{\circ}$. } \label{fig:pde}
\end{figure*}

We concur from the above that finding the horizontal, vertical and
diagonal Haar coefficients after some rotation reduces to the problem of applying the chain rule to the
coefficients of the Haar transform of $f$ prior to rotation - achieving, thus, rotation of the coefficients directly in the transform domain. Again, this is true because Haar coefficients conveniently correspond to function derivatives at different scales. Figure \ref{fig:pde} shows an
example of computing $\frac{\mbox{\small{d\textsl{g}}}}{\mbox{\small{d}}\phi}$ after a
vertical elevation of $\alpha_l=20^{\circ}$, where
$\frac{\partial{f}}{\partial\Theta}$ and
$\frac{\partial{f}}{\partial\Phi}$ are obtained from the vertical
coefficients $\frac{\partial{f}}{\partial\theta}$ and the horizontal
coefficients $\frac{\partial{f}}{\partial\phi}$, respectively, through
an elevation by the same angle.

This provides an elegant analytic solution to the rotation of Haar wavelets defined over a spherical coordinate system. Furthermore, our method can be applied to any map other than the longitude-latitude map as long as there exists a mathematical expression that maps the data from the sphere to all pixels of a rectangular image. In other words, if $\Theta(\theta,\phi)$ and $\Phi(\theta,\phi)$ can be explicitly expressed mathematically then our method for rotation can be used for spherical data represented using that map.

Taking advantage of the longitude-latitude map, one can reduce the complexity of the rotation matrix $\textbf{R}$. Assuming that a rotation by $\theta_l$ along the elevation angle is aligned with the rotation around the X-axis, then $\theta_l$ equals $\alpha_l$. This in turn means that a rotation by $\theta_l$ can be represented by $\textbf{R}_x$. A rotation with respect to the azimuth angle by $\phi_l$ simply becomes a linear shift of the elevated point. This reduces $\textbf{R}$ from a multiplication of the three euler rotation matrices to one matrix and a linear shift. This significantly reduces the number of cosine and sine terms in $\textbf{R}$ and, therefore, reducing its complexity.

\subsection{MATHEMATICAL DESCRIPTION}

The basic idea here is that the Haar coefficients at any level $j+1$ of a function $f$ defined over the spherical coordinates can be considered as the horizontal, vertical and the diagonal derivatives of $f$ with respect to the elevation and azimuth angles at that level (scale). Therefore, given the Haar coefficients of $f$ at level $j+1$, we can directly compute the derivatives $g_\theta(\theta,\phi)=\frac{\mbox{d}g}{\mbox{d}\theta}$, $g_\phi(\theta,\phi)=\frac{\mbox{d}g}{\mbox{d}\phi}$, and $g_{\theta\phi}(\theta,\phi)=\frac{\mbox{d}^2g}{\mbox{d}\phi\mbox{d}\theta}$ of the rotated function $g$ at level $j+1$ by simply using the equations (\ref{eq:pde1})-(\ref{eq:pde3}). To obtain the Haar coefficients of $g$ at level $j$, we simply need to convolve these derivatives with $2\times 2$ averaging kernels and then downsample the results.

Simple inspection then shows that Haar coefficient at all coarser levels $j-1, j-2 , j-3, ...$ can be computed directly from the derivatives $g_\theta$, $g_\phi$, and $g_{\theta\phi}$ by a series of convolutions and downsampling. We describe the main idea using the vertical coefficients as an example. Suppose we want to compute the vertical coefficients $v_{j-l}$ at level $j-l$ using the vertical derivatives $g_\theta$. This amounts to convolving the vertical derivatives with two separate kernels as follows:
\begin{equation}
v_{j-l}(\theta,\phi)=\sum_\rho h_{s,j-l}(\rho-\phi)\sum_\tau g_\theta(\theta,\phi)h_{t,j-l}(\tau-\theta) \label{eq:conv}
\end{equation}

where
\begin{equation}
h_{t,j-l}(\theta)=\left(2^{l+1}-|\theta|\right) \;\;\; , \;\;\; \theta \in [-2^{l+1}+1,2^{l+1}-1] \label{eq:tker}
\end{equation}
and
\begin{equation}
h_{s,j-l}(\phi)=\frac{1}{2^{l+2}}\;\;\; , \;\;\; \phi \in [-2^{l+1}+1,2^{l+1}-1] \label{eq:sker}
\end{equation}
followed by downsampling by a factor of $2^{l+2}$.

Horizontal Haar coefficients at level $j-l$ can be similarly computed from $g_\phi$ by interchanging the role of $h_t$ and $h_s$. The diagonal coefficients are simply computed by taking the cross-derivative either using the horizontal or the vertical coefficients and convolving in both directions by $h_s$.

\subsection{ALGORITHMIC OPTIMIZATION}

The simple but powerful results derived in the previous section show that we can directly compute the Haar coefficients of a rotated function at all levels $j-l$, $l=0,1,2,...$ directly from the Haar coefficients of the original function at only one level, i.e. level $j+1$. However, it can be verified from (\ref{eq:conv})-(\ref{eq:sker}) that as $l$ increases the time complexity of (\ref{eq:conv}) increases exponentially. At a first glance, this may seem disappointing, however, it turns out that this is rather misleading, and there is a simple solution to reduce the complexity to provide run-time computation.

There are two measures that we can take to optimize the algorithm: First, note that the Haar coefficients at all levels $j-l$, $l=0,1,2,...$ are computed from the derivatives at level $j+1$. Second the convolution kernels $h_{t,j-l}$ and $h_{s,j-l}$ increase exponentially in size as $l$ increases. Both problems can be alleviated by identifying the fact that the proposed computations in the previous section can be performed recursively so that at each level the computation of the Haar coefficients depends only on the Haar coefficients computed at the previous level. To this end, we note that for all $l\geq 1$, the kernel $h_{t,j-l}$ can be written as the convolution of two kernels:
\begin{equation}
h_{t,j-l}=h_{t,j}*h^l_{t,j}
\end{equation}
where
\begin{equation}
h^l_{t,j}(\theta)=\left\{\begin{array}{ll}
h_{t,j}\left(\frac{\theta}{2^{l}}\right) & \mbox{if $\frac{\theta}{2^{l}}$ is an integer}\\
0 & \mbox{otherwise}
\end{array}\right.
\end{equation}
Note that except for three values, all the values in the kernel $h^l_{t,j}$ are zero. Essentially, $h^l_{t,j}$ is the same kernel as $h_{t,j}$, but upsampled by a factor of $2^{l}$, by zero padding all elements between the non-zero values.  Therefore, the algorithm can be implemented recursively by repeated convolution with the upsampled kernel $h^l_{t,j}$ for all $l\geq 1$. The latter is very cheap since except for three values all the the values in the kernel are zero. Simple inspection indicates that the reduced algorithm is $\emph{O}(n)$.

In addition to the above algorithmic optimization for recursive
computation of the Haar coefficients, we can further reduce the
computational time by modifying also the rotation step based on the
chain rule described in Section \ref{sec:rot}. To this end note that
without loss of generality, we can assume that the function to be
rotated is initially aligned such that the $x$-axis coincide with
the $\theta$-axis. The consequence of this assumption is that all
rotations reduce to rotation around the $\theta$-axis (or the
$x$-axis) followed by a simple shifting along the $\phi$-axis, which
is extremely cheap (i.e. $\emph{O}(1)$).

The following algorithm summarizes our method:
\begin{itemize}
\item For each pixel, determine the local outgoing direction and retrieve the BRDF transformed map corresponding to that direction.
\item Assuming the transformed BRDF data has $0...n-1$ levels of resolution, the vertical, horizontal and diagonal coefficients at level $n-1$ are rotated by $\phi_{\vec{N}}$ and $\theta_{\vec{N}}$, which are the azimuth and elevation angles of the normal $\vec{N}$ at the current pixel.
\item The coefficients at the lower resolution levels are subsequently evaluated by recursively convolving by the filters $h_s$ and $h_t$, where $h_s$ and $h_t$ have a size of 2 and 3 coefficients respectively.
\item The rotated data is plugged in the triple integral computation.
\end{itemize}

The above algorithm has a complexity of $\emph{O}(N)$, where N is the number of coefficients. One might argue that this is the same complexity as saving multiple levels of resolution of untransformed data using a mipmap \cite{Lee09}, rotating the required level spatially and then transforming. Representing the data using a wavelet transform, however, is more compact than a mipmap, which saves storage space and bandwidth as is our goal. It is also more convenient to have the data in its transformed state so that they are readily available for use for the triple product computation.

We would also like to mention that one can start at any resolution level that is lower than $n-1$. In that case, the computational complexity is reduced to $\emph{O}(N/4^k)$, where $k$ is the number of levels removed.

\section{EXPERIMENTAL RESULTS AND DISCUSSION}
Our algorithm is implemented using CUDA ("Compute Unified Device Architecture") in combination with Cg for graphics rendering. The algorithm is implemented into the following passes:

\textbf{The Cg Pass}: This pass takes advantage of the interpolator and rasterizer of the graphic pipeline to output a fragment buffer that contains the interpolation information, vertex id, and object id per pixel.

\textbf{The Rotation Input Pass}: This pass outputs a buffer which contains the BRDF tile id and the rotation angles required per non-background pixel.

\textbf{The Rotation Pass}: This pass is actually divided into three passes that run concurrently to optimize speed. The three passes evaluate the horizontal, vertical and diagonal coefficients of the rotated BRDF tiles.

\textbf{The Triple Integral Pass}: This pass evaluates the illumination of each non-background pixel using the triple sum to render a 3D object in a scene based on the light captured by images.

One immediate question is whether the third step of run-time rotation in the Haar domain has an effect in the accuracy of the rendering of the object, and if so to what extent? To answer this question, we processed the BRDF data by rotating it in the spatial domain to generate the ground truth. We then processed the BRDF data by rotating it using our method of run-time rotation in the Haar domain. To make sure that we gather statistically meaningful information, we performed this process using five different popular models (``sphere'', ``Easter'', ``Venus'', ``bunny'', and ``cat'') and rendered them for two different types of surface material (aluminium bronze and blue metallic paint) using five different environment maps, i.e. ``Galileo Tomb'', ``Grace Cathedral'', ``RNL'' (light probe), ``St. Petersburg'', and ''Uffizi'' gallery in Florence. We performed this over a large set of random rotations at levels 5 and 6, and measured the average PSNR. Results are shown in Tables \ref{tab:tab1}-\ref{tab:tab4}. It can be immediately noticed that generally speaking these PSNR values are extremely high in the range of 100dB-120dB. It is worth noting that, for instance, in a typical state of the art lossy compression the PSNR is in the range 30dB-50dB. This indicates that the accuracy of run-time rotation in the Haar domain is on average 2-3 times better than the best available lossy compression. A quick inspection of these tables also suggest that the rotation accuracy decreases when performed at coarser levels of Haar, but not drastically so if we do not start at a too coarse level. Therefore, the gain in speed may justify the slight loss of accuracy up to some level of coarseness.
\begin{table}[h!]
\caption{Average PSNR for rendering different geometric objects made of aluminium bronze and rotated randomly at level 6 within different environment maps.}\label{tab:tab1}
\begin{center}
{\scriptsize
\begin{tabular}{|c|c|c|c|c|c|}
  \hline
  Environment Map & Galileo & Grace & RNL & St. Petersburg & Uffizi \\ \hline
  PSNR (dB) & 110.82 & 110.92 & 111.54 & 110.76 & 111.02 \\
  \hline
\end{tabular}
}
\end{center}
\vspace*{0.5cm}
\caption{Average PSNR for rendering different geometric objects made of aluminium bronze and rotated randomly at level 5 within different environment maps.}\label{tab:tab2}
\begin{center}
{\scriptsize
\begin{tabular}{|c|c|c|c|c|c|}
  \hline
  Environment Map & Galileo & Grace & RNL & St. Petersburg & Uffizi \\ \hline
  PSNR (dB) & 111.17 & 111.81 & 112.56 & 111.29 & 111.62 \\
  \hline
\end{tabular}
}
\end{center}
\vspace*{0.5cm}
\caption{Average PSNR for rendering different geometric objects made of blue metallic paint and rotated randomly at level 6 within different environment maps.}\label{tab:tab3}
\begin{center}
{\scriptsize
\begin{tabular}{|c|c|c|c|c|c|}
  \hline
  Environment Map & Galileo & Grace & RNL & St. Petersburg & Uffizi \\ \hline
  PSNR (dB) & 104.28 & 104.64 & 104.97 & 104.66 & 104.73 \\
  \hline
\end{tabular}
}
\end{center}
\vspace*{0.5cm}
\caption{Average PSNR for rendering different geometric objects made of blue metallic paint and rotated randomly at level 5 within different environment maps.}\label{tab:tab4}
\begin{center}
{\scriptsize
\begin{tabular}{|c|c|c|c|c|c|}
  \hline
  Environment Map & Galileo & Grace & RNL & St. Petersburg & Uffizi \\ \hline
  PSNR (dB) & 105.49 & 106.11 & 106.20 & 105.55 & 105.49 \\
  \hline
\end{tabular}
}
\end{center}
\end{table}

\begin{figure*}[t!]
\begin{center}
\begin{tabular}{cc}
\includegraphics[width=3.5in,height=2.5in]{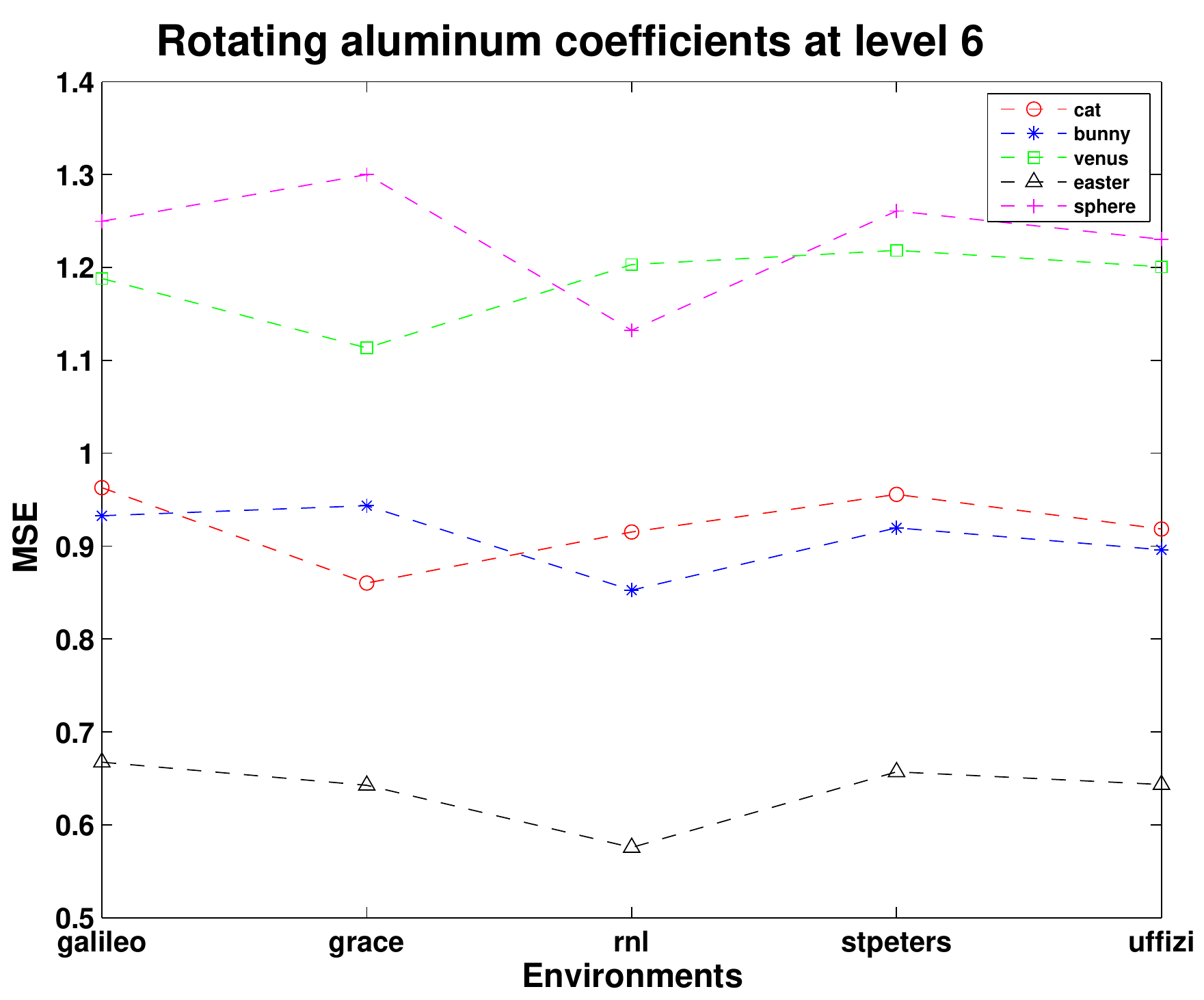} &
\includegraphics[width=3.5in,height=2.5in]{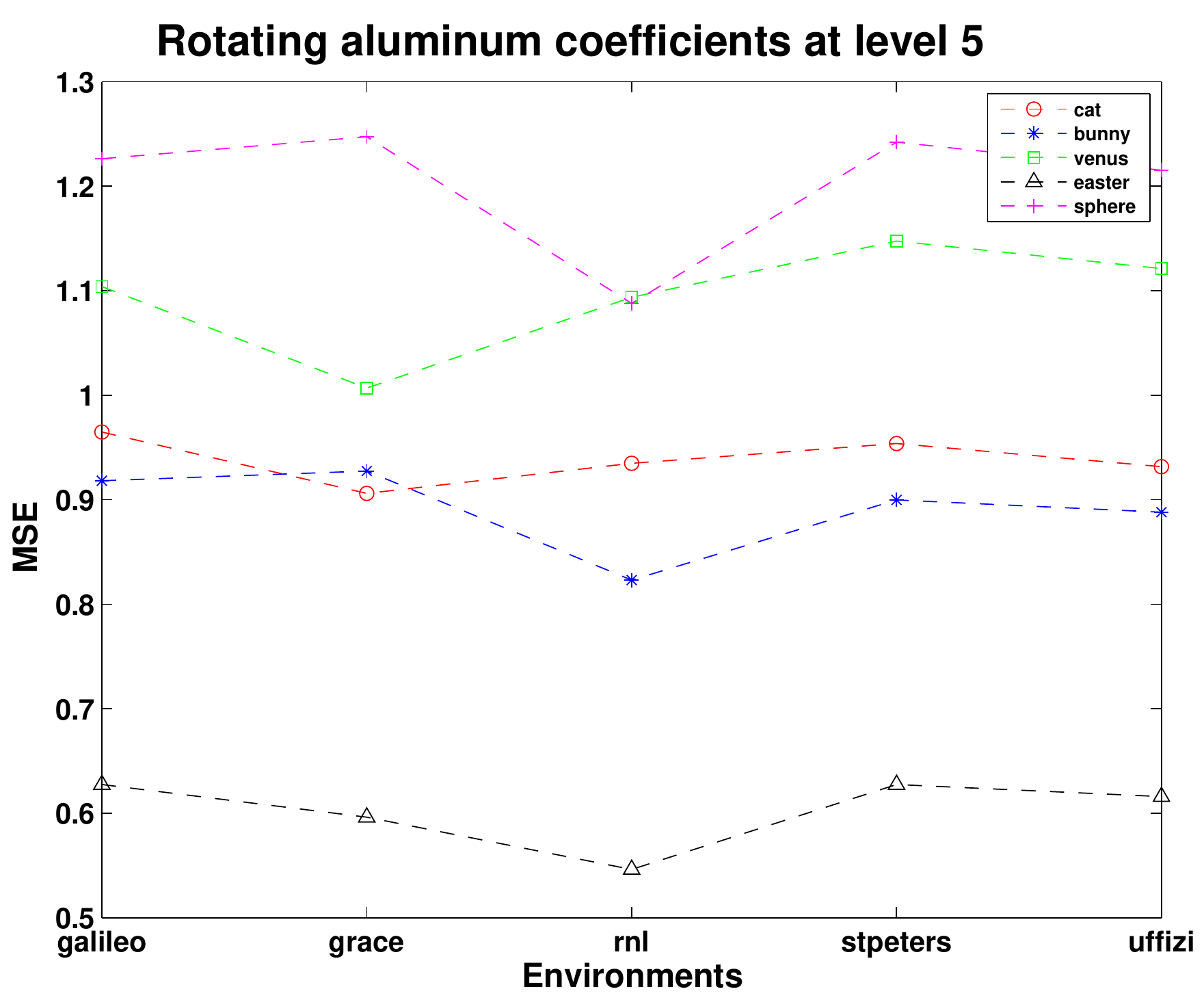}
\end{tabular}
\end{center}
\caption{The plots show the MSE, comparing rendered images of different models using aluminum bronze material under different environments. The errors are computed by comparing preprocessed data under spatial rotation with our method of rotation. Using our method, we rotated the coefficients at level 6 then evaluated the coefficients at the coarser levels, recursively.} \label{fig:alumError}

\begin{center}
\begin{tabular}{cc}
\includegraphics[width=3.5in,height=2.5in]{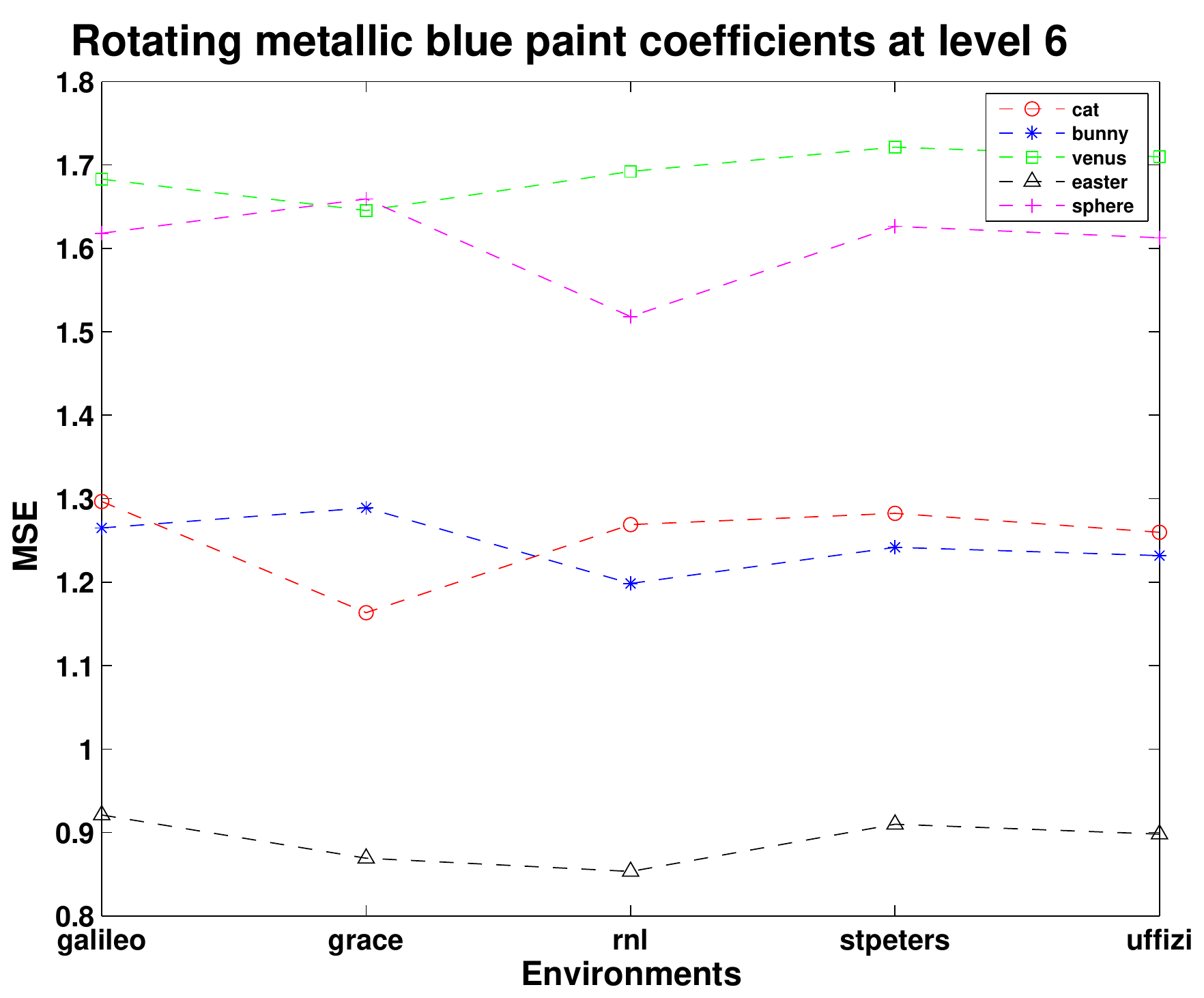} &
\includegraphics[width=3.5in,height=2.5in]{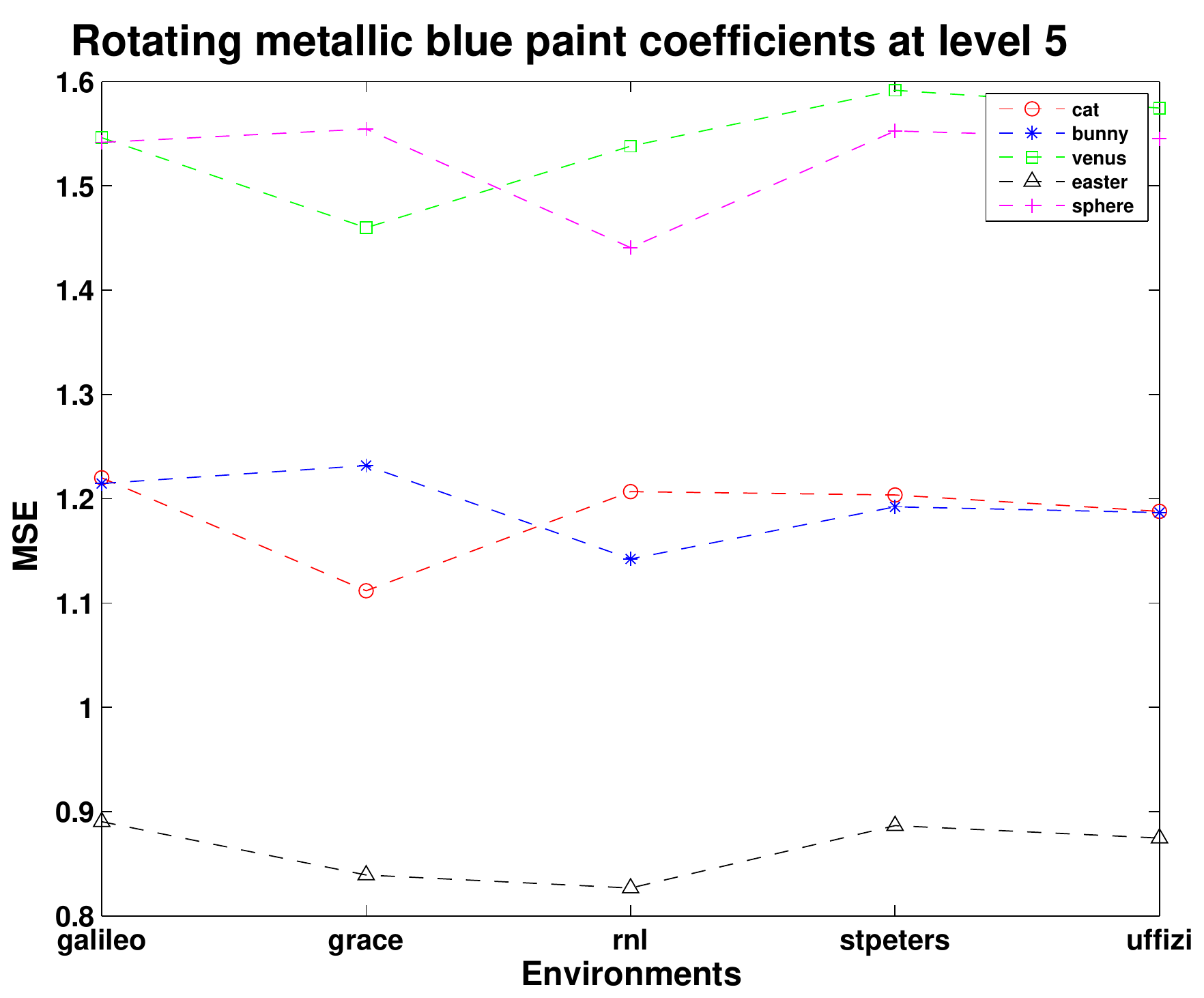}
\end{tabular}
\end{center}
\caption{The plots show the MSE, comparing rendered images of different models using blue metallic paint material under different environments. The errors are computed by comparing preprocessed data under spatial rotation with our method of rotation. Using our method, we rotated the coefficients at level 6 then evaluated the coefficients at the coarser levels, recursively.} \label{fig:blueError}
\vspace*{0.5cm}
\begin{center}
\begin{tabular}{cc}
\includegraphics[width=3.4in,height=2.2in]{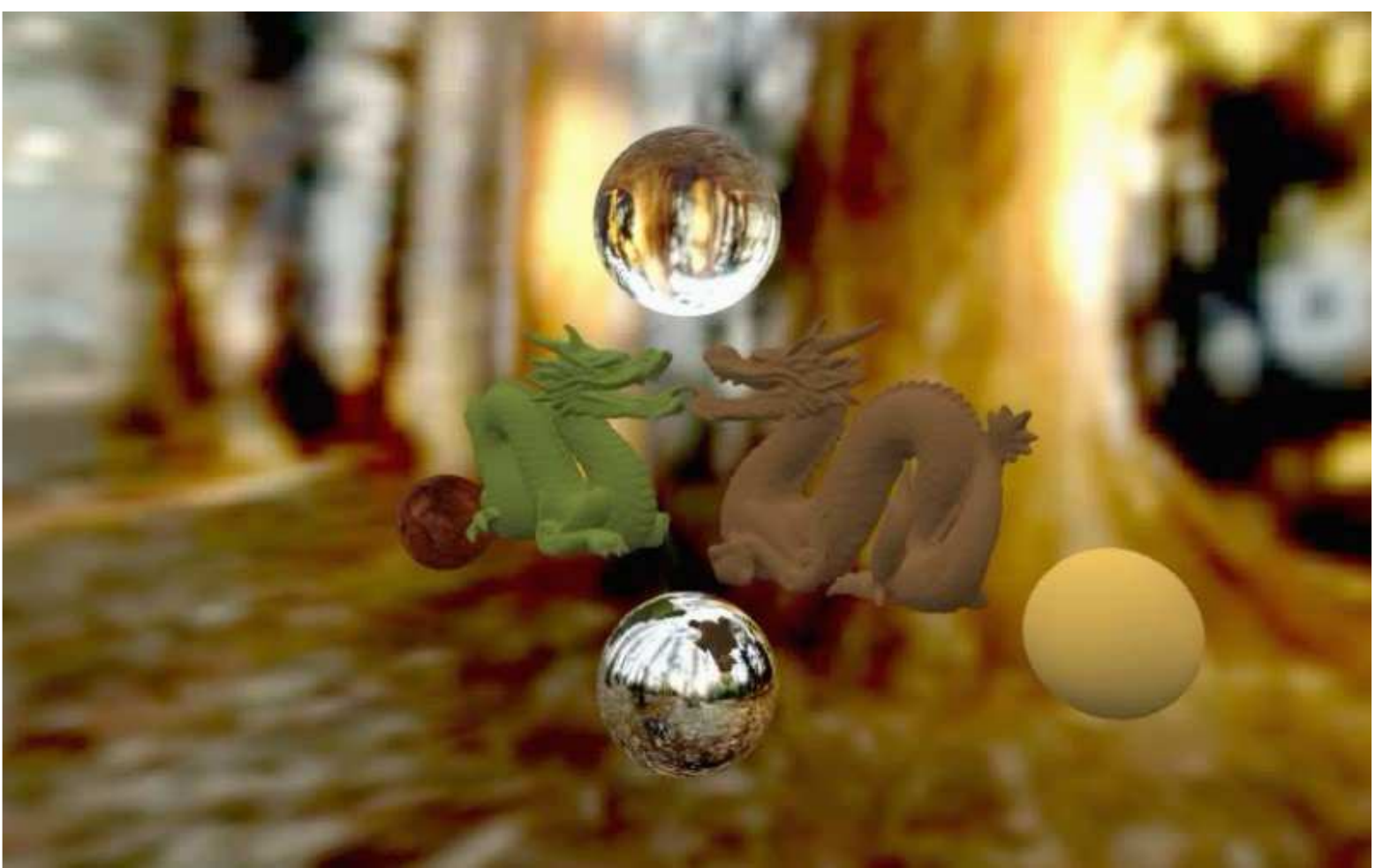} &
\includegraphics[width=3.4in,height=2.2in]{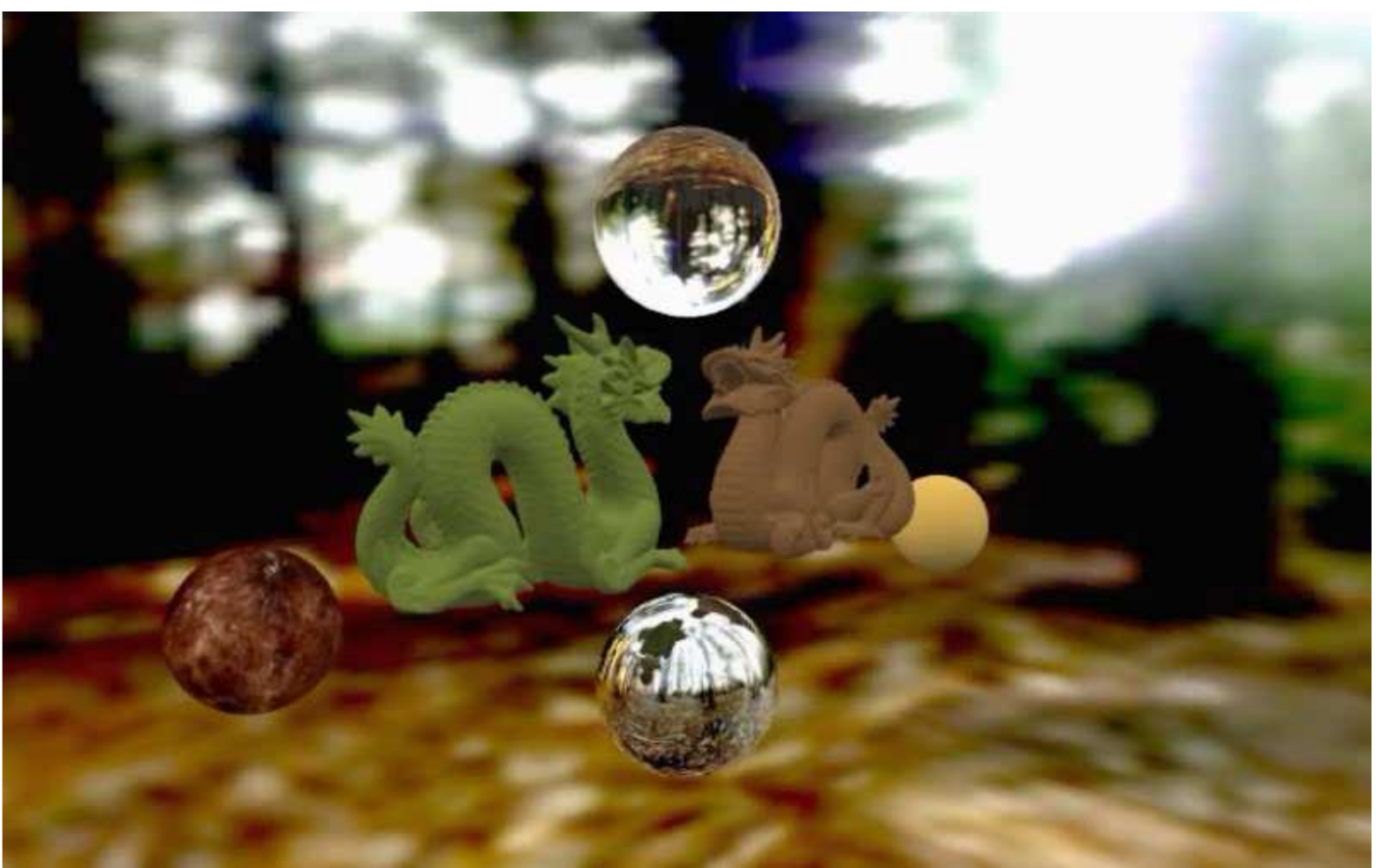}
\end{tabular}
\end{center}
\caption{Objects with different levels of glossiness (low and high frequency reflectance) and transparency rotated and rendered within the same environment map.} \label{fig:scene2}
\end{figure*}

\begin{figure*}[h]
\begin{center}
\begin{tabular}{ccc}
\includegraphics[width=2.2in]{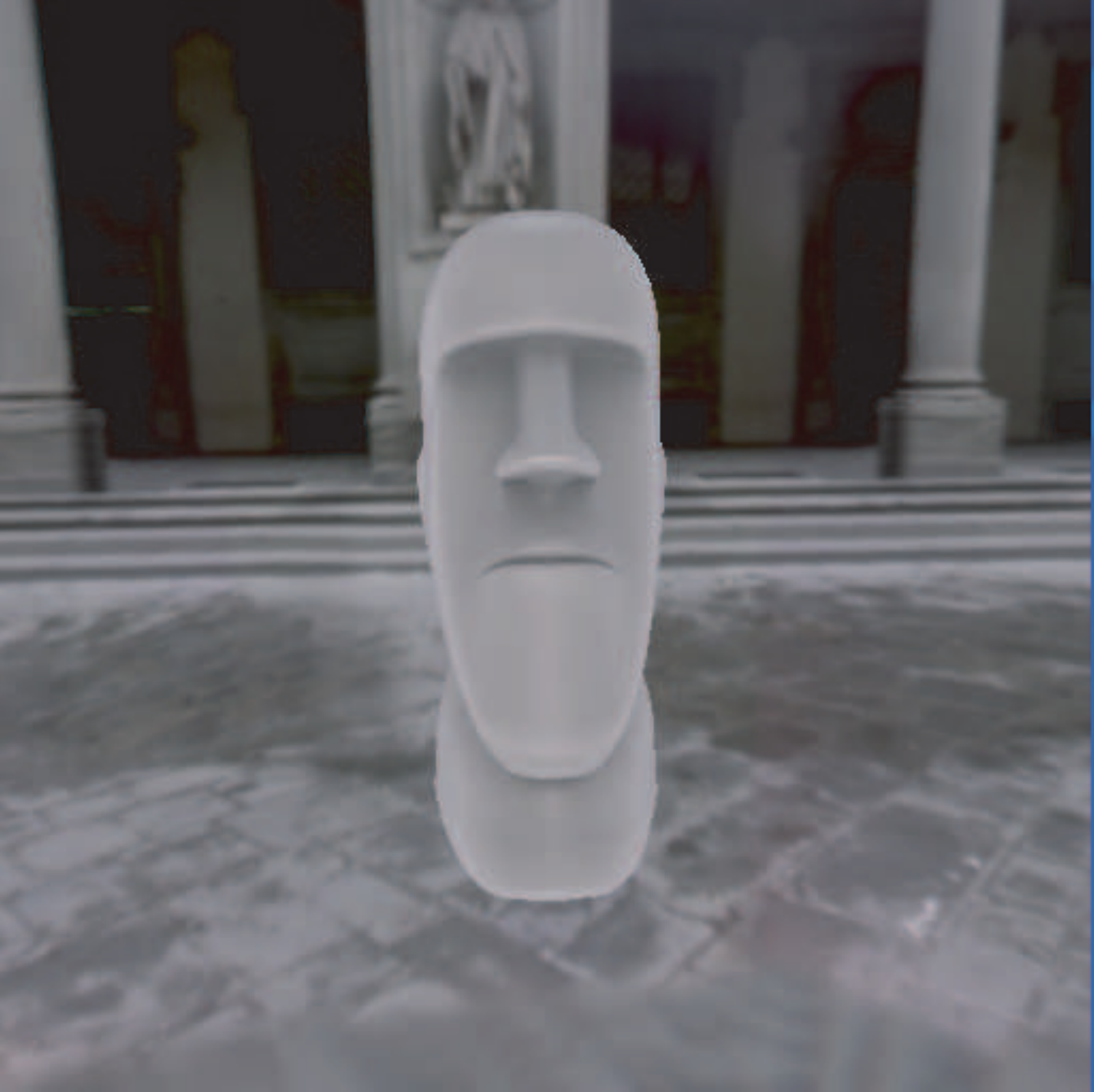} &
\includegraphics[width=2.2in]{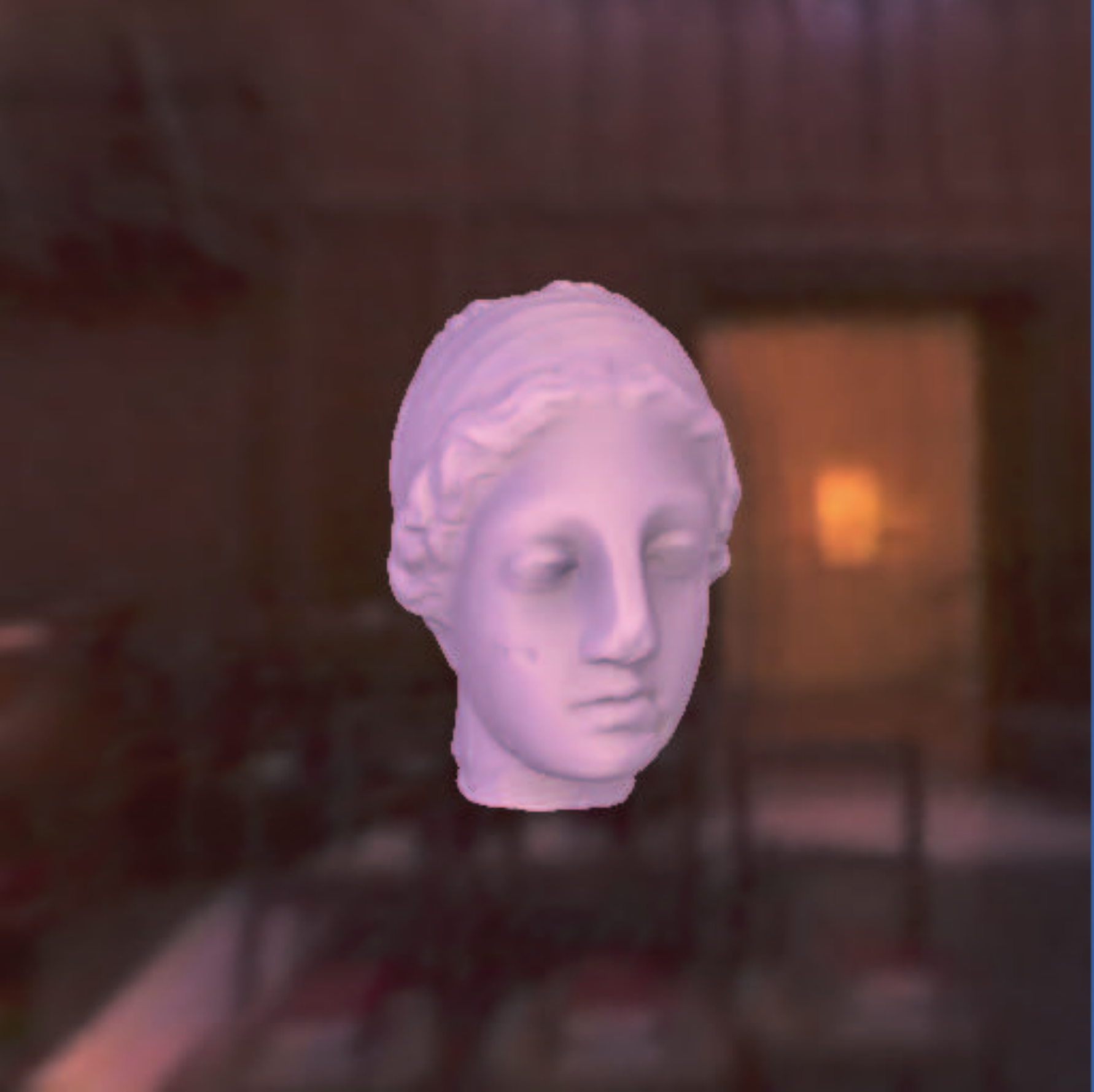} &
\includegraphics[width=2.2in]{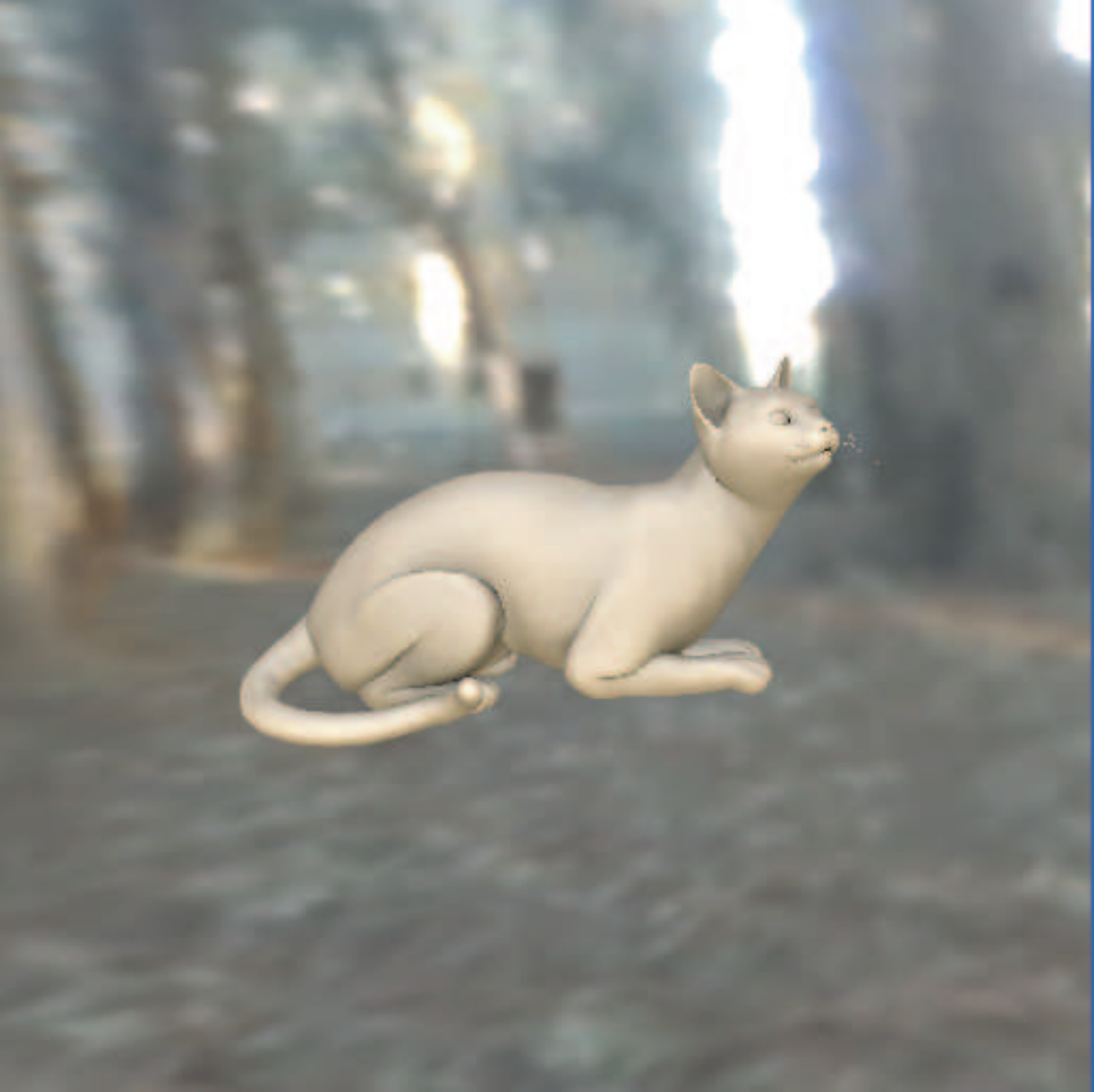} \\
\includegraphics[width=2.2in]{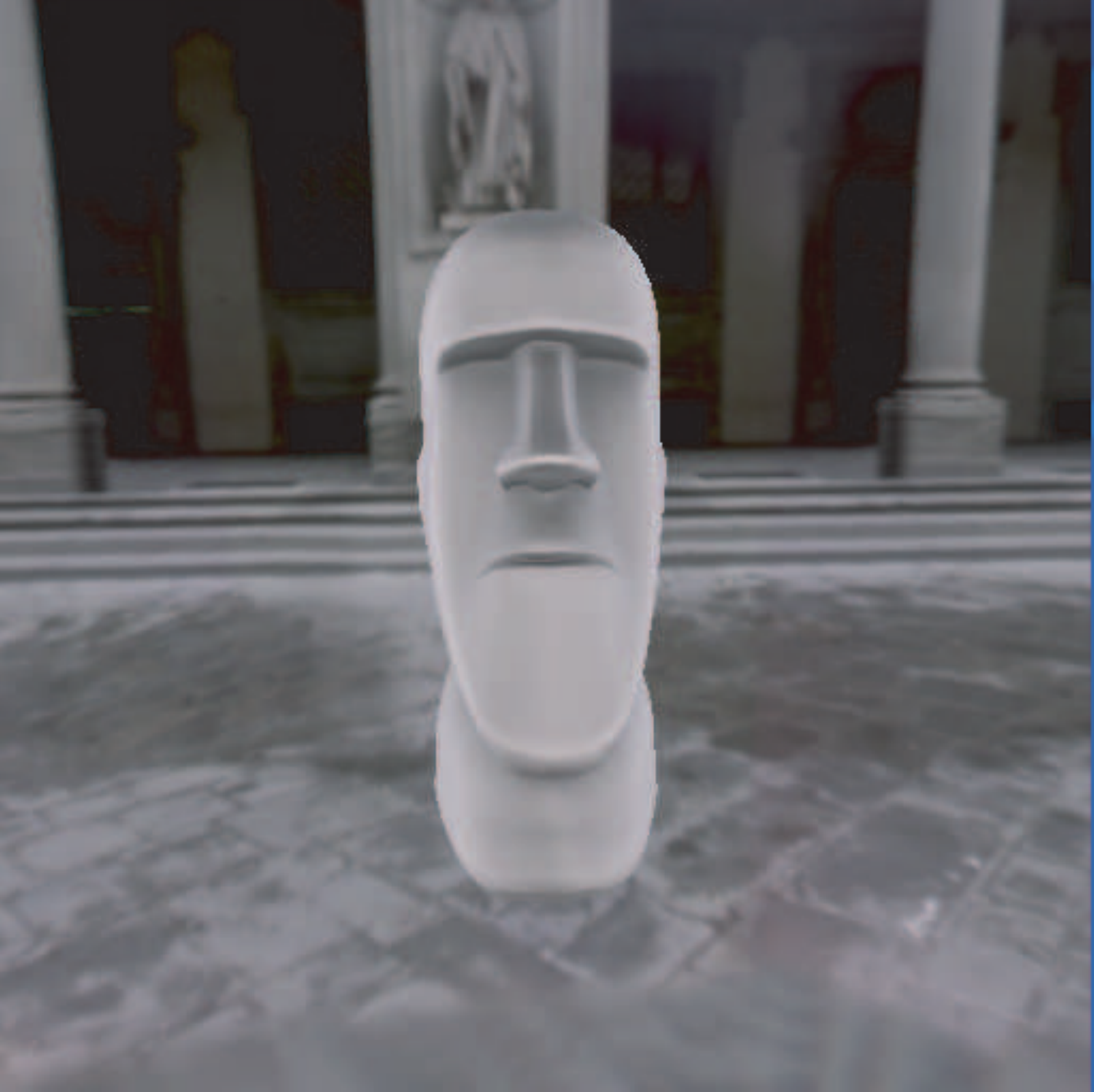} &
\includegraphics[width=2.2in]{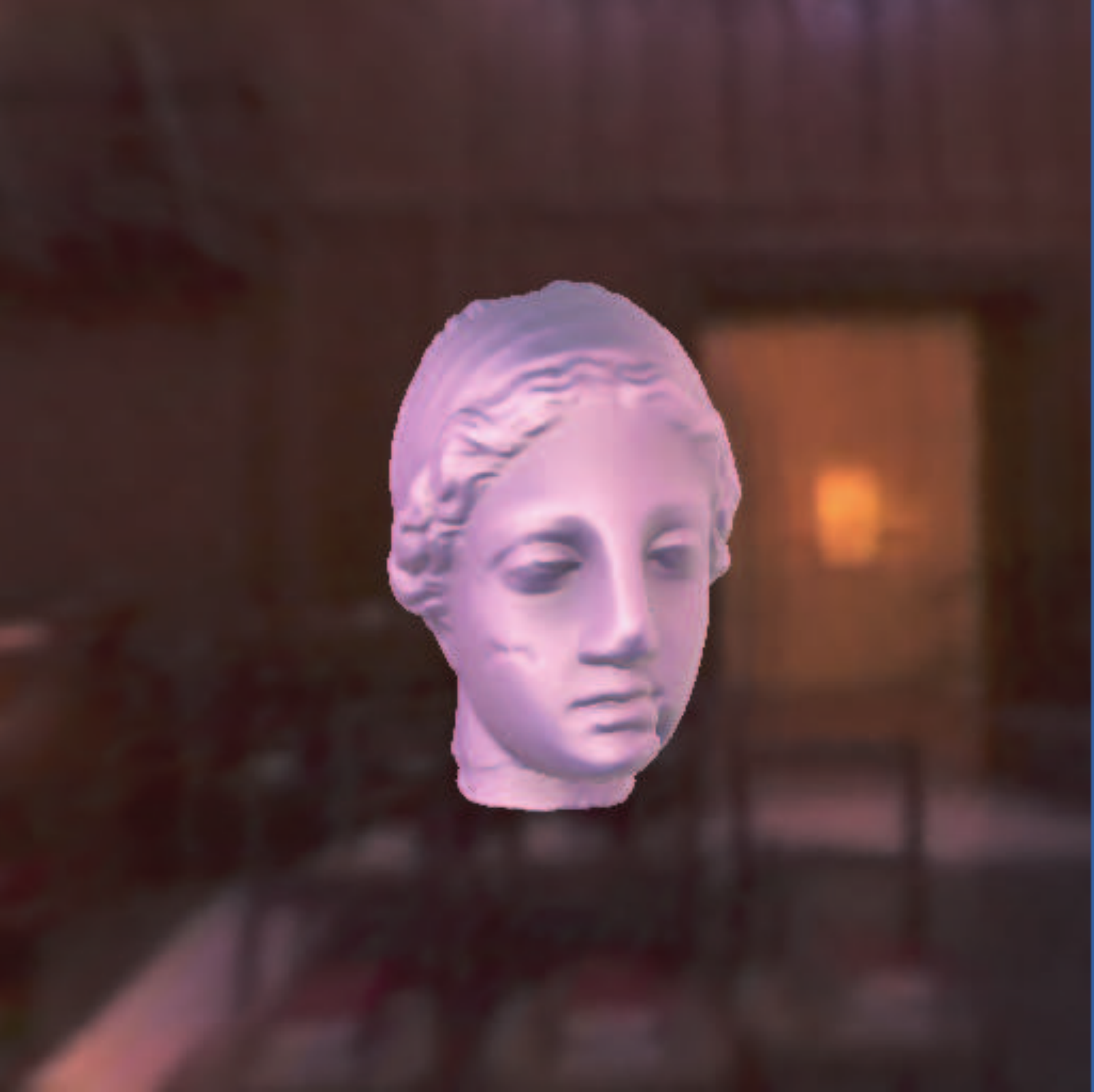} &
\includegraphics[width=2.2in]{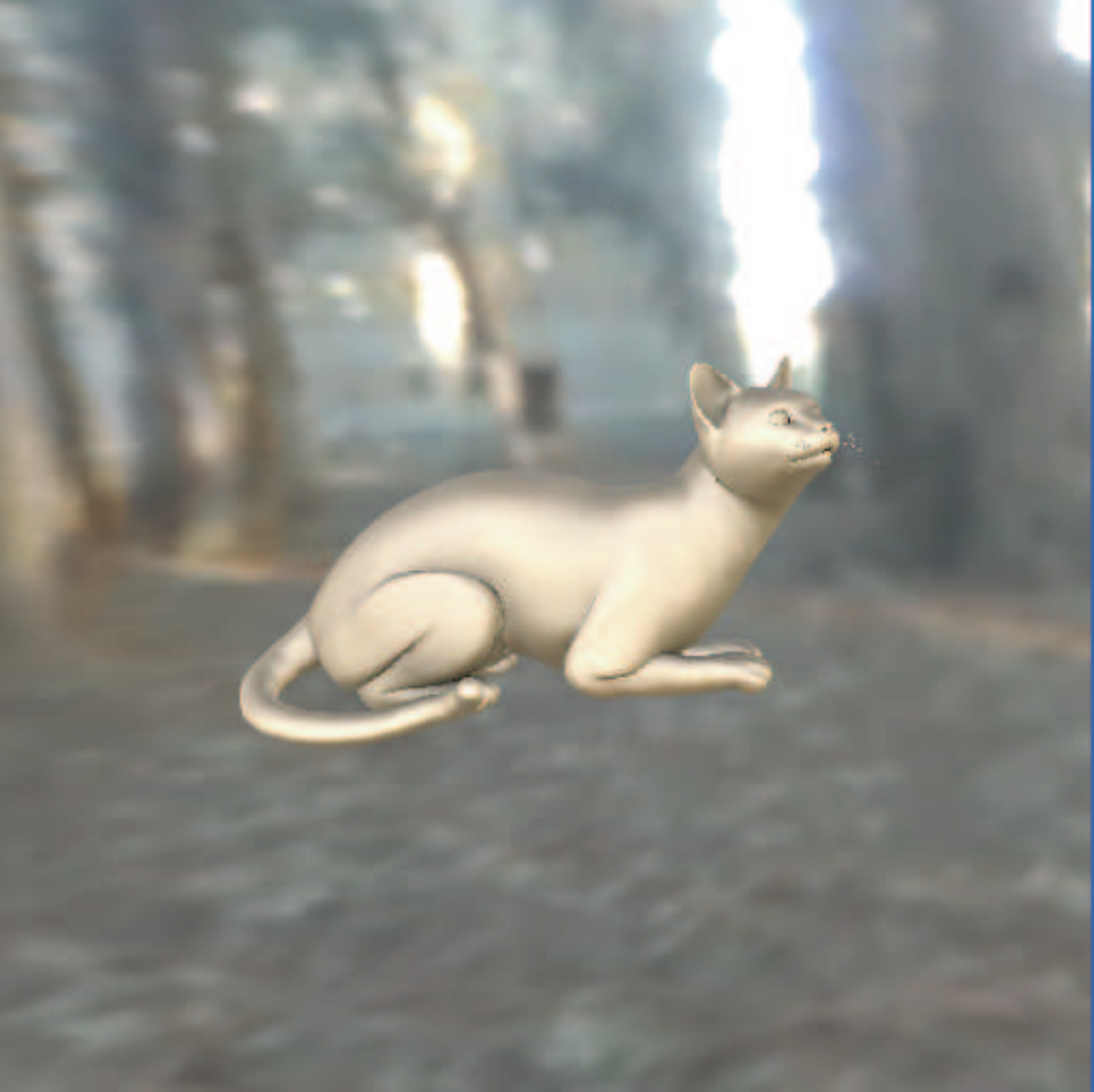}  \\
\includegraphics[width=2.2in]{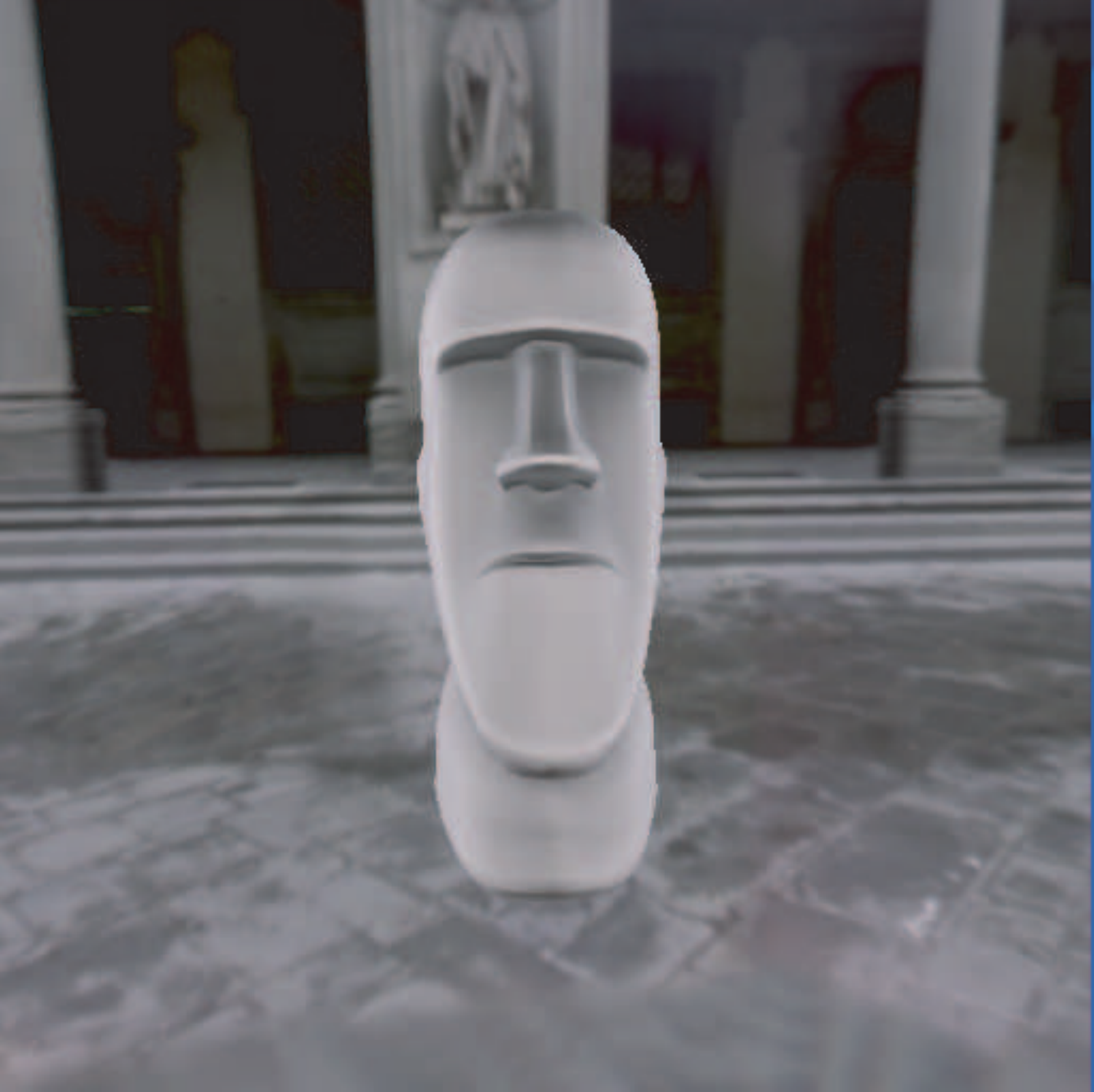} &
\includegraphics[width=2.2in]{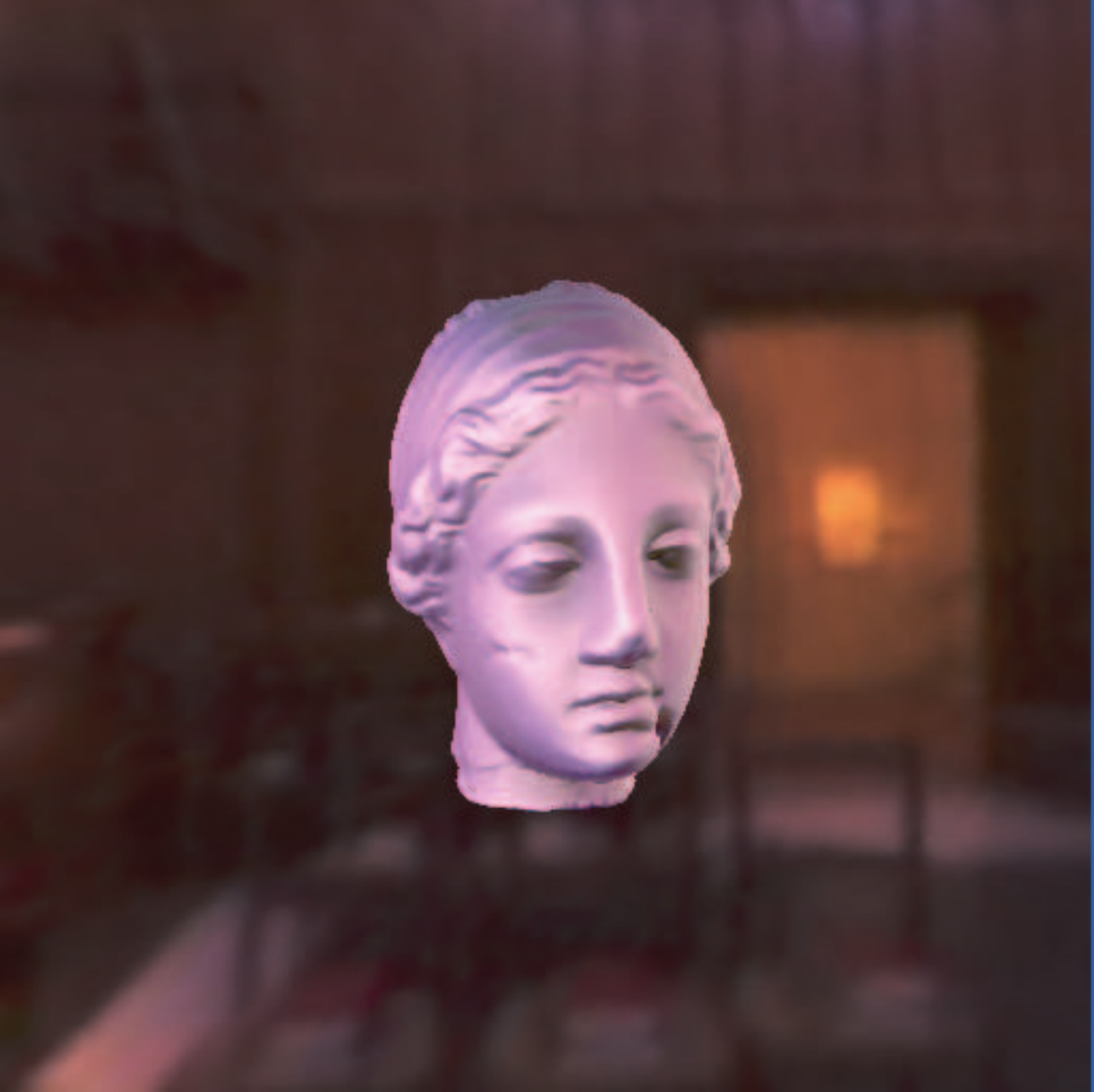} &
\includegraphics[width=2.2in]{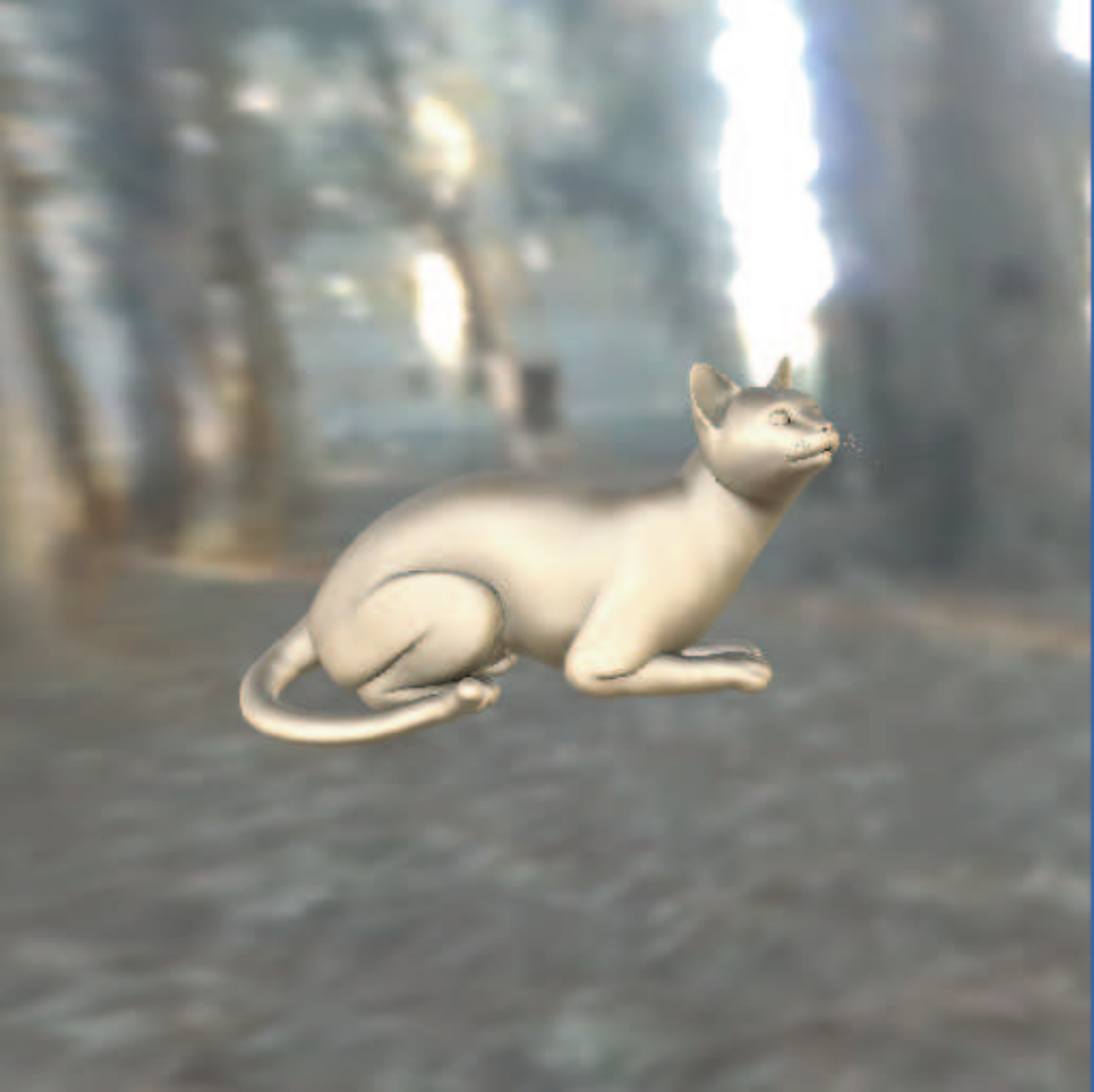} \\
\includegraphics[width=2.2in]{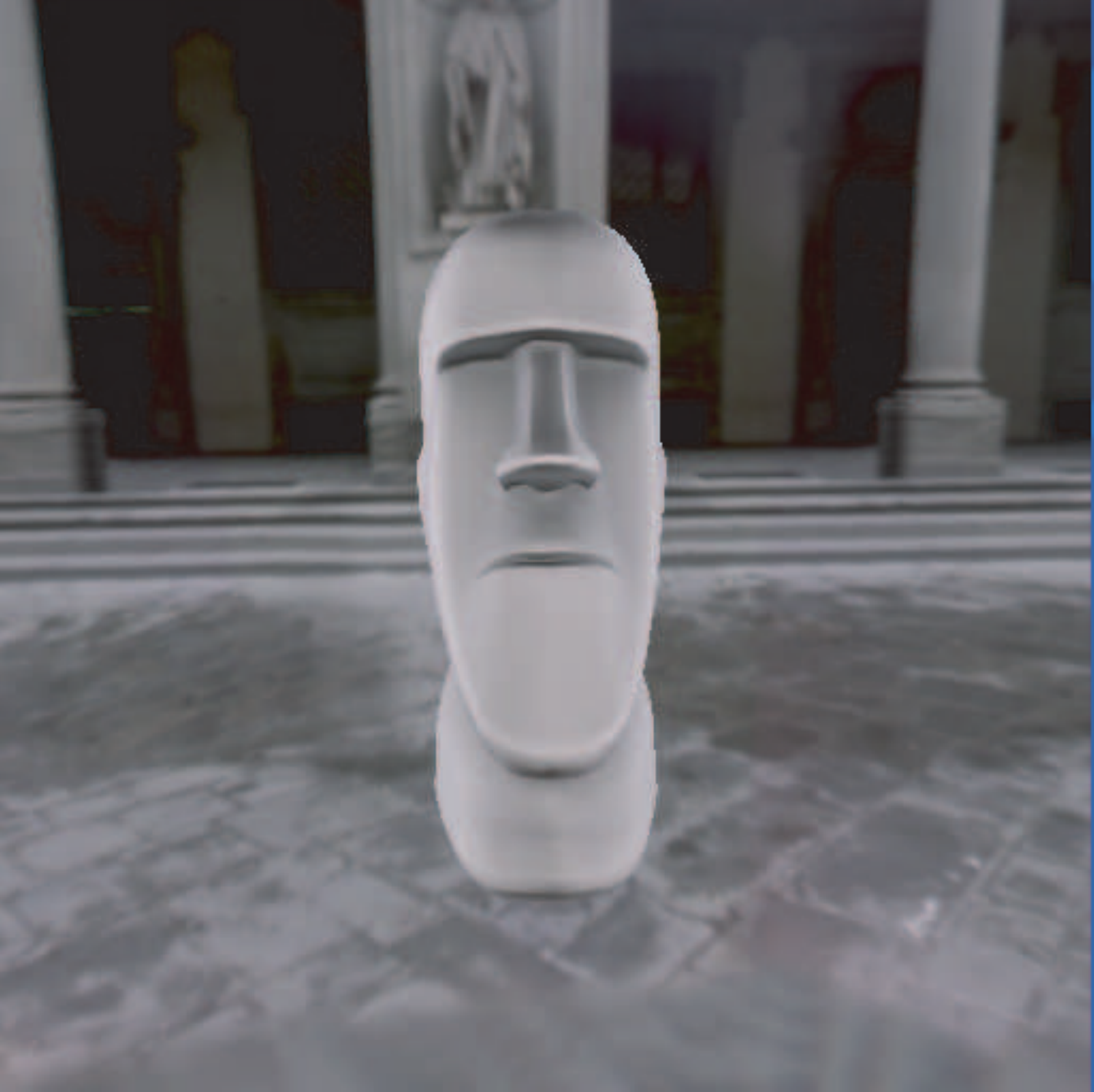} &
\includegraphics[width=2.2in]{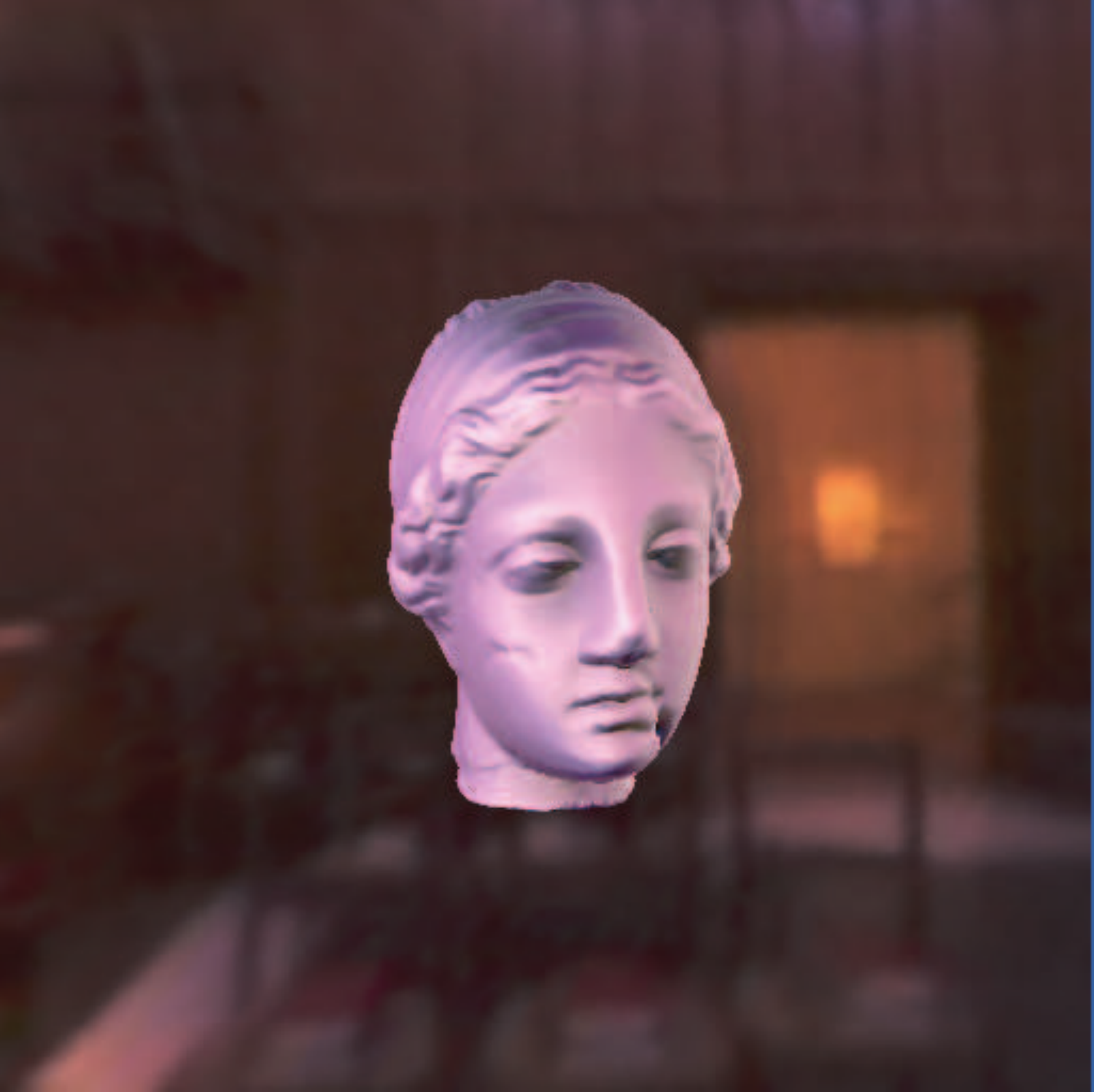} &
\includegraphics[width=2.2in]{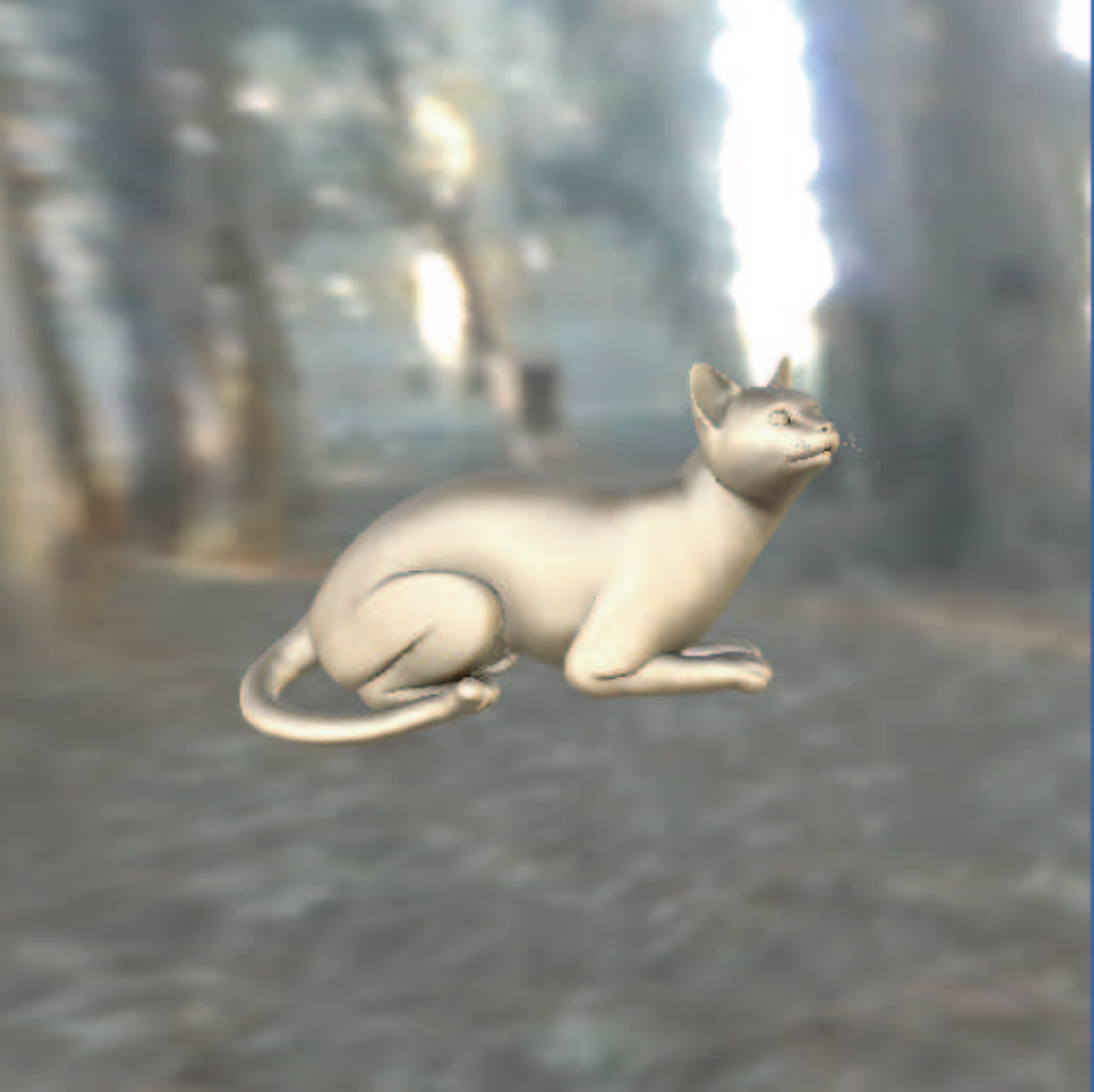}
\end{tabular}
\end{center}
\caption{Different degrees of glossiness. The rows of images are rendered with 4, 16, 64 and 256 coefficients, respectively. The three columns are rendered using steel, blue metallic paint and aluminum bronze, respectively. } \label{fig:shininess}
\end{figure*}

In order to also evaluate the level of errors in terms of the different object geometry, we computed the mean square error (MSE) averaged over a large set of random rotations for the same models, materials, and environment maps as above. The average MSE is very close to zero (see the plots in Figures \ref{fig:alumError} and \ref{fig:blueError}). The figures show results from rotating the data from levels 5 and 6 and recursively evaluating the rest of the coefficients. Rotating from level 5 is slightly better because it has less accumulation error because of the smaller number of levels. The smaller the number of coefficients at a certain level the less smooth the data becomes, which causes inaccuracy due to discontinuity. Also, a simple inspection of these plots show that the MSE remains invariant to the environment map (lighting variations). This is significant because it implies that under dynamic lighting the MSE would remain invariant, and hence would not create visually disturbing noise.   
We then experimented with rendering different levels glossiness as a function of the bandwidth, i.e. the number of Haar coefficients used to render a specific material. Fig. \ref{fig:scene2} shows an example of objects with different levels of glossiness (low and high frequency lighting) and transparency rotated and rendered together within the same environment map. Figure \ref{fig:shininess} shows experimenting with three different materials, i.e. steel, blue metallic paint and aluminum bronze, as we increase the number of Haar coefficients used from 4 to 256. These experiments show that by using a fairly small number of Haar coefficients our method can produce variable levels of glossiness. Therefore, since we do not store rotated BRDF's or rotation matrices, our method is essentially a compressed-domain processing method for rendering materials at different levels of high-frequency content.

Our most important contribution is providing the first real solution to rotating Haar wavelets. Although one method of rotation already exits \cite{Wang06WaveRot}, it does so by creating a rotation matrix per discretized rotation angle. Each rotation matrix is created during a preprocess stage where each wavelet of a resolution less than or equal the required image size is rotated spatially then transformed and stored as one column of that specific rotation matrix. The rotation matrices are then used during rendering time to rotate any of the required data. This solution is computational rather than analytical, therefore, it relies on preprocessing and discretization, which produce errors due to interpolation.

We are currently able to rotate coefficients at level 4 using the GPU and at level 5 using the CPU. We recursively generate the coefficients at the coarser levels on the GPU. This is the equivalent of using 256 and 1024 coefficients for rendering respectively. Although PRT methods are capable of linearly compressing data and generating glossier materials at the current time, they still suffer from excess storage, discretization and interpolation error. It is a well-known fact that the speed of computing systems increase at a higher rate than memory, especially now that the market is focusing on parallel computing both on the CPU and GPU. In a few years time our method will be able to render using a larger number of coefficients and achieve interactive/real time.

\section{CONCLUSION}

In this paper, we aspired to establish the grounds for our work
by giving a brief but necessary introduction to wavelets, their
different properties and the state of wavelet related research on
the property of shift-invariance. Achieving shift-invariance was,
up until now, the only method to achieve phase-shifting. However,
that compromised other desired properties, which we wish to keep
for the purposes of the environment lighting application at hand.
These properties, namely, orthogonality, perfect reconstruction
and localization are essential to solving the triple product that
results from projecting the lighting integral terms into frequency
space and, therefore, are a necessity to preserve. We present our
current work for phase-shifting, which does not rely on
shift-invariance and therefore preserves these important
properties. Our recent work on linear phase-shifting of Haar wavelets was demonstrated in \cite{Alnasser2D}. In this paper, we discussed the environment lighting
problem and its relationship to Haar wavelets. Our goal is to
achieve realistic image-based rendering and relighting of synthetic objects in a scene using Haar
wavelets as the integration medium, while improving the efficiency
in terms of storage over other methods by means of rotating during rendering time.
In order to do so, we devised a novel method to tackle the
non-linear phase-shifting that makes use
of the fact that the Haar transform implicitly contains the
horizontal, vertical and diagonal derivatives of the signal. Our run-time radiance transfer method (RRT) provides an elegant solution to the non-trivial problem of rotating Haar wavelets, providing thus a solution for run-time computation of light transport based on the rendering equation. The resulting algorithm scales nicely so that no trade-off is required in terms of storage, computational cost, and bandwidth. Furthermore, since the proposed solution does not require precomputation, errors due to interpolation are avoided. Finally, our approach may be viewed as the first compressed-domain rendering method for PIBR, since all other methods in this area are storage and bandwidth intensive.

%
%
%
%
%
%
\clearpage

\bibliographystyle{plain}
\bibliography{foroosh,mais}

\end{document}